\documentclass[]{bytedance_seed}

\usepackage[toc,page,header]{appendix}

\usepackage{natbib}

\usepackage{CJKutf8}

\usepackage{xargs}  

\usepackage{todonotes}  

\usepackage{multirow}

\usepackage{cleveref}

\usepackage{amsmath}
\usepackage{dsfont}

\usepackage{subcaption}

\usepackage{svg}

\usepackage{mathrsfs}
\usepackage{adjustbox}
\usepackage{multirow}
\usepackage{multicol}
\usepackage{tcolorbox}
\usepackage{changepage}
\usepackage{enumitem}
\usepackage{graphicx}
\usepackage{amssymb}
\usepackage{xcolor}
\usepackage{tabularx} 
\usepackage{makecell} 
\usepackage[dvipsnames]{xcolor}

\usepackage{minitoc}

\title{ RewardDance: Reward Scaling in Visual Generation}

\author{
  Jie Wu\texorpdfstring{$^{\ast,\dagger}$}{} \hspace{0.1cm}
  Yu Gao\texorpdfstring{$^{\ast}$}{} \hspace{0.1cm}
  Zilyu Ye \hspace{0.1cm}
  Ming Li \hspace{0.1cm}
  Liang Li \hspace{0.1cm}
  Hanzhong Guo  \hspace{0.1cm} 
  Jie Liu \\[0.1cm]
  Zeyue Xue \hspace{0.1cm}
  Xiaoxia Hou \hspace{0.1cm}
  Wei Liu \hspace{0.1cm}
  Yan Zeng \hspace{0.1cm}
  Weilin Huang\texorpdfstring{$^{\ddagger}$}{}
}

\affiliation[]{ByteDance Seed}
\contribution[*]{Equal contribution} 
\contribution[\ddagger]{Corresponding authors} 
\contribution[\dagger]{Project lead}

\begin{document}
\begin{CJK*}{UTF8}{gbsn}

\abstract{

Reward Models (RMs) are critical for improving generation models via Reinforcement Learning (RL), yet the RM scaling paradigm in visual generation remains largely unexplored. It primarily due to fundamental limitations in existing approaches: CLIP-based RMs suffer from architectural and input modality constraints, while prevalent Bradley-Terry losses are fundamentally misaligned with the next-token prediction mechanism of Vision-Language Models (VLMs), hindering effective scaling. More critically, the RLHF optimization process is plagued by Reward Hacking issue, where models exploit flaws in the reward signal without improving true quality.
To address these challenges, we introduce \textbf{RewardDance}, a scalable reward modeling framework that overcomes these barriers through a novel generative reward paradigm. 
By reformulating the reward score as the model's probability of predicting a "yes" token, indicating that the generated image outperforms a reference image according to specific criteria – RewardDance intrinsically aligns reward objectives with VLM architectures. This alignment unlocks scaling across two dimensions: (1) \textit{Model Scaling}: Systematic scaling of RMs up to 26 billion parameters; (2) \textit{Context Scaling}: Integration of task-specific instructions, reference examples, and chain-of-thought (CoT) reasoning.  Extensive experiments demonstrate that RewardDance significantly surpasses state-of-the-art methods in text-to-image, text-to-video, and image-to-video generation. Crucially, we resolve the persistent challenge of "reward hacking": Our large-scale RMs exhibit and maintain high reward variance during RL fine-tuning, proving their resistance to hacking and ability to produce diverse, high-quality outputs. It greatly relieves the mode collapse problem that plagues smaller models.
}

\maketitle

\begin{figure}[h]
\begin{center}
\vspace{-30pt}
\includegraphics[width=0.88\linewidth]{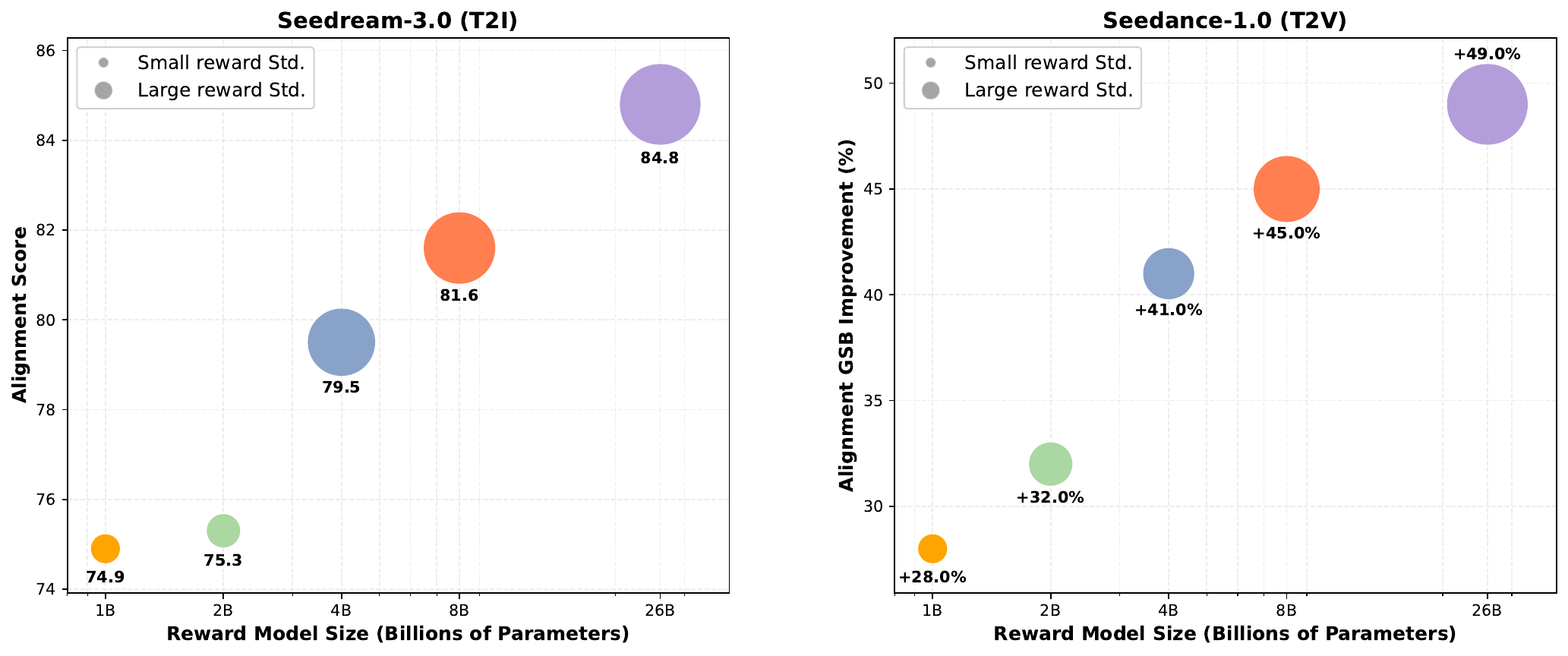}
\end{center}
\vspace{-6pt}
\caption{
\scriptsize RewardDance consistently boosts T2I and T2V generation (validated by alignment/GSB). The reward variance during the later stages of RL training, represented by bubble size, serves as an indicator of policy "hacking." Low variance implies mode collapse, where the model tends to generate uniform, high-reward outputs. High variance signifies the policy maintains output diversity across various prompts, indicating that it has not collapsed.
}
\vspace{-20pt}
\label{fig:teaser}
\end{figure}


\newpage



\section{Introduction}

Diffusion models have achieved immense progress in visual generation. State-of-the-art models such as FLUX~\cite{blackforestlabs_flux} and Seedream~\cite{gong2025seedream,gao2025seedream} for image generation, alongside Wan2.1~\cite{wan2025wan} and Seedance~\cite{gao2025seedance} for video generation, have unlocked vast creative potential. These capabilities are further enhanced by paradigms like Reinforcement Learning (RL)~\cite{wallace2024diffusion,dong2023raft,fan2023reinforcement,gupta2025simple,xue2025dancegrpo,liu2025flow,controlnet++,superedit,multireward} and Test-time Scaling~\cite{ma2025inference}, where the Reward Model (RM) plays a pivotal role. While a robust and accurate RM can significantly improve generation quality, the community has lacked clear guidance on designing superior RMs. In this paper, we introduce RewardDance, a new framework built on the principle that \textit{scalability is the key to creating better visual RMs}.

\begin{table}[h]
\centering
\resizebox{\textwidth}{!}{%
\begin{tabular}{c|c|c|c|c|c|ccc}
\toprule
\multirow{2}{*}{\textbf{Methods}} & \multirow{2}{*}{\textbf{Task}} & \multirow{2}{*}{\textbf{Aligning Stage}} & \multirow{2}{*}{\textbf{Base Model}} & \multirow{2}{*}{\textbf{Modeling Paradigm}} & \multirow{2}{*}{\textbf{Model Scaling}} & \multicolumn{3}{c}{\textbf{Context Scaling}} \\
             &       &      &     &        &   & Task Instruction & Reference Examples & CoT Data \\
\hline
ImageReward \cite{xu2023imagereward}  & Visual & RL & CLIP & Regressive & $\times$ & $\times$  & $\times$    & $\times$        \\
PickScore \cite{kirstain2023pick}    & Visual& RL& CLIP & Regressive & $\times$ & $\times$  & $\times$    & $\times$        \\
HPSv2 \cite{wu2023human}        & Visual& RL& CLIP & Regressive & $\times$ & $\times$  & $\times$    & $\times$        \\
VisionReward \cite{xu2024visionreward} & Visual& RL& VLM  & Regressive & $\times$ & $\checkmark$  & $\times$    & $\times$        \\
VideoAlign \cite{liu2025improving}   & Visual& RL& VLM  & Regressive & $\times$ & $\checkmark$  & $\times$    & $\times$        \\
HPSv3 \cite{ma2025hpsv3}  & Visual& RL\&Infer & VLM  & Regressive  & $\checkmark$ & $\checkmark$  & $\times$    & $\times$ \\
WorldPM \cite{wang2025worldpm}  & Understanding& RL & VLM  & Regressive  & $\checkmark$ & $\checkmark$  & $\times$    & $\times$ \\
DeepSeek-GRM \cite{liu2025inference}  & Understanding& Infer & VLM  & Generative  & $\times$ & $\checkmark$  & $\checkmark$    & $\times$ \\
Pairwise RM \cite{xu2025unified} & Understanding& RL & VLM  & Generative  & $\times$ & $\checkmark$  & $\times$    & $\times$ \\
UnifiedReward \cite{wang2025unified}  & Multimodal& - & VLM  & Generative  & $\times$ & $\checkmark$  & $\checkmark$    & $\checkmark$ \\
RewardDance  & Visual& RL\&Infer & VLM  & Generative  & $\checkmark$ & $\checkmark$  & $\checkmark$    & $\checkmark$ \\

\bottomrule
\end{tabular}%
}
\caption{A comprehensive comparison of visual and multimodal reward models. Our RewardDance is the first framework for visual generation to successfully integrate a generative paradigm with comprehensive scalability across both reward model size and reward context dimensions (task instructions, reference examples, and CoT data).}
\label{tab:comparison}
\end{table}

\begin{figure}[hp]
  \centering
  \includegraphics[width=1.0\textwidth, keepaspectratio]{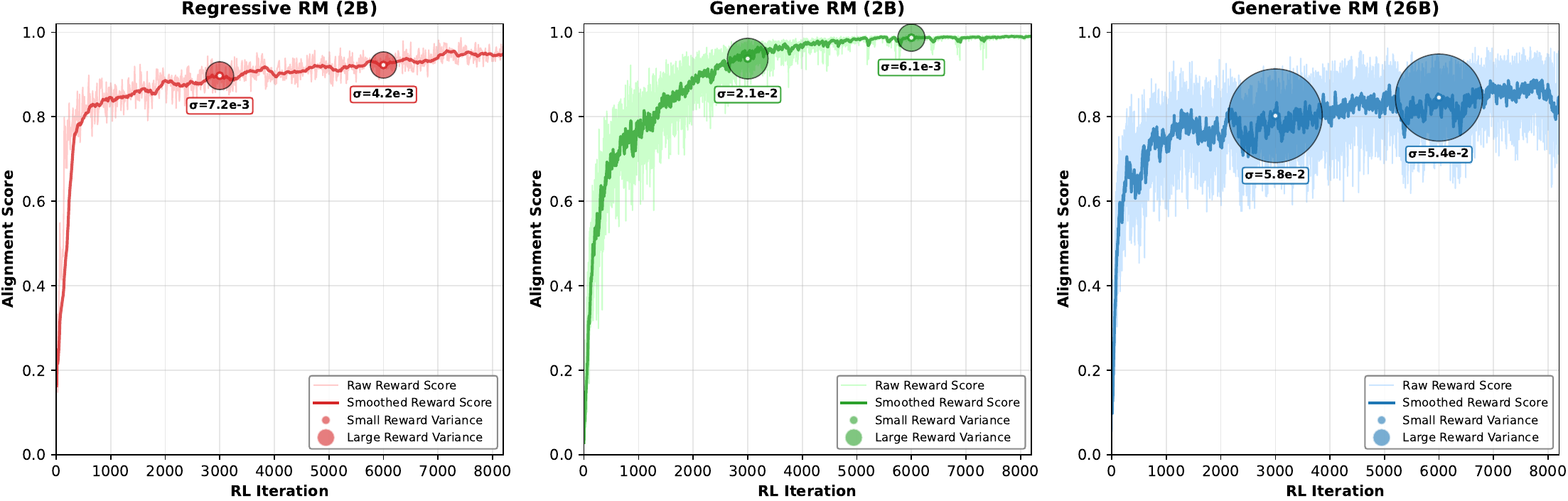}
  \vspace{-5mm}
  \caption{Comparison of training dynamics for Regressive vs. Generative reward models during diffusion RL fine-tuning. At the same 2B model scale (Left vs. Middle panel), the generative reward model exhibits significantly superior training dynamics compared to the regression-based one: it facilitates higher exploration magnitude, manifested as greater reward variance, and a more favorable reward growth trend. This higher diversity in reward signals indicates that the generative RM exhibits stronger robustness against reward hacking. Under a regression-based RM, the diffusion model risks learning to exploit reward loopholes to achieve high scores without making substantive progress. This inherent robustness is key to the generative RM's successful scaling to 26B parameters (Right panel).}
  \label{fig:regressive-generative-rm-comparison}
\end{figure}

Our work is motivated by a comprehensive analysis of existing reward modeling methods, as summarized in Table~\ref{tab:comparison}. Early CLIP-based reward models~\cite{xu2023imagereward, wu2023human} were limited by CLIP’s architecture, which was difficult to scale~\cite{li2023inverse} and generalized poorly to diverse tasks~\cite{li2024exploring}.
Later VLM-based models explored new paradigms and scaling strategies, but progress has been fragmented: some achieve large-scale models yet remain limited to regressive paradigms (e.g., HPSv3 \cite{ma2025hpsv3} , WorldPM~\cite{wang2025worldpm}), while others adopt stronger generative paradigms without effective scaling (e.g., UnifiedReward~\cite{wang2025unified}). As shown in the Figure \ref{fig:regressive-generative-rm-comparison} , the regression-based reward model is very susceptible to reward hacking.

We introduce RewardDance to address above challenges. It represents the first framework designed to achieve this unification, leveraging a generative paradigm that enables effective scaling across both model size and context—including task-specific instructions, reference examples, and Chain-of-Thought (CoT) reasoning—to unlock the full potential of VLMs for advanced visual reward modeling.

Our RewardDance framework resolves this fundamental paradigm mismatch by rethinking reward modeling as a generative task. Instead of appending a regression head, we convert the reward score into the VLM's predicted probability of a "yes" token. This approach natively aligns the reward objective with the VLM's autoregressive, next-token prediction mechanism, thereby enabling effective reward scaling along two primary dimensions:

\begin{itemize}[leftmargin=*]

\item \textbf{Model Scaling:} We break from the conventional practice of using smaller, fixed-size models by systematically scaling our RM with VLMs of increasing capacity (1B to 26B parameters). This directly links model parameter count to both reward modeling performance and final generation quality.

\item \textbf{Context Scaling:} In contrast to traditional methods that rely on simple image-text pairs, RewardDance enriches input context by incorporating task-aware instructions, reference examples, and CoT reasoning data, enabling more robust and accurate reward judgments.

\end{itemize}

To validate RewardDance, we conducted extensive experiments on both open-source and proprietary models across a range of tasks, including text-to-image, text-to-video, and image-to-video generation. We observed that as the parameter of the Reward Model (RM) increases, the diffusion model exhibits stronger exploratory capacity at higher RLHF iteration steps and is less susceptible to reward hacking. Moreover, incorporating additional context information significantly enhances model performance and robustness. Crucially, our findings reveal a pronounced scaling effect: under both RL and test-time scaling paradigms, visual generation quality consistently and stably improves in direct correspondence with reward model enhancement.

Our contributions are summarized as:
\begin{itemize}[leftmargin=*]
\item \textbf{Scalability as the Principle for RMs:} We establish scalability as the fundamental design principle for visual RMs, addressing a critical yet underexplored dimension in existing work and providing new insights for the field.
More importantly, our experiments demonstrate that scaling up the Reward Model (RM) enables less reward hacking (Fig.~\ref{fig:regressive-generative-rm-comparison}) and further achieves significant improvements in generation quality (Fig.~\ref{fig:teaser}).

\item \textbf{Generative Reward Modeling:} We propose a novel generative reward modeling paradigm that resolves the fundamental mismatch in prior work by regarding reward prediction as a next-token prediction task. This naturally aligns with VLMs' autogressive mechanism, unlocking the potential for effective reward scaling.

\item \textbf{Comprehensive Reward Scaling:} We propose and validate a holistic methodology for scaling RMs across two dimensions: reward model size (from 1B to 26B parameters) and reward context (including task-aware instructions, reference examples, and CoT reasoning). We are the first to systematically demonstrate that enhancing these dimensions yields stable and consistent scaling effects, where improved reward models directly contribute to higher-quality visual generation.
\end{itemize}

\section{Related Work}
\subsection{Diffusion Models}
Generalized diffusion models encompass both traditional diffusion processes~\cite{sohl2015deep,ho2020denoising,song2020score} and flow-based models~\cite{lipman2022flow}. These methods learn to map a simple prior distribution, typically a Gaussian, to a target data distribution by reversing a learned noise-injection process. Owing to their effectiveness in modeling complex, high-dimensional data, diffusion models have demonstrated remarkable performance across a wide range of tasks, including image~\cite{rombach2022high,podell2023sdxl,esser2024scaling,gong2025seedream,blackforestlabs_flux,gao2025seedream,liao2025mogao,zhang2024unifl} and video~\cite{blattmann2023stable,guo2023animatediff,brooks2024video,bao2024vidu,gao2025seedance,seawead2025seaweed,zeng2024make,kong2024hunyuanvideo,wan2025wan,ma2025step,zhang2024onlinevpo,chen2023control} generation.

\subsection{Reward Models}
Reward Models (RMs) play a pivotal role in aligning the outputs of visual generation models with human preferences~\cite{ren2024byteedit,zhang2024unifl,gao2025seedance,gong2025seedream,gao2025seedream}. Early approaches, such as ImageReward~\cite{xu2023imagereward}, PickScore~\cite{kirstain2023pick}, and HPSv2~\cite{wu2023human}, fine-tuned CLIP models to produce scores that reflect human preferences. Subsequent works~\cite{xu2024visionreward,liu2025improving} adopted Vision-Language Models (VLMs) as backbones, typically appending a regression head to output reward signals. These methods, predominantly trained with the Bradley–Terry loss~\cite{bradley1952rank,ouyang2022training,ziegler2019fine}, are categorized as regression-based RMs. More recently, advancements in LLMs and multimodal learning have inspired new paradigms. For instance, WorldPM~\cite{wang2025worldpm} investigated the scaling potential of RMs through enhanced data and model architectures. Concurrently, a new class of generative reward models was proposed~\cite{liu2025inference,xu2025unified,wang2025unified}, aiming to better leverage the native capabilities of pre-trained models. Building upon these developments, our work introduces the generative RM paradigm to the visual domain and, for the first time, systematically explores its scaling properties with respect to both reward model size and reward context.

\subsection{Reinforcement Learning from Human Feedback}

Reinforcement Learning from Human Feedback (RLHF) has been increasingly adopted in diffusion models to align generated outputs with human preferences. The typical pipeline involves training a reward model on human preference data and leveraging it to guide the generative model. For instance, DDPO~\cite{black2023training} adapts Proximal Policy Optimization (PPO) to diffusion frameworks by computing image log-likelihoods. In contrast, ReFL~\cite{xu2023imagereward} circumvents likelihood computation challenges by directly optimizing diffusion outputs through gradients from a frozen reward model. Beyond diffusion models, recent works~\cite{ye2025schedule,xue2025dancegrpo,liu2025flow,liu2025improving} have extended RLHF techniques to flow-based generative models, demonstrating the broader applicability of preference-based optimization in generative modeling.
Complementing training-time alignment methods, test-time optimization techniques have emerged to enhance diffusion model performance during inference without requiring model retraining. These approaches~\cite{ma2025inference,oshima2025inference,liu2025improving} typically involve reward-guided adjustments to noise prediction outputs or dynamic modifications to sampling schedules, thereby improving generation quality in a plug-and-play manner. In this work, we integrate our RewardDance framework into both reinforcement learning and inference-time optimization stages, demonstrating its effectiveness in aligning generated content with human preferences across diverse visual generation tasks.
\section{Method}

This section introduces RewardDance, our framework for scalable visual reward modeling. We first detail its core generative paradigm, then describe our methodologies for scaling it along two primary dimensions: context and model size. Finally, we outline the training pipeline in aligning diffusion models.

\subsection{Preliminary}
As illustrated in Figure~\ref{fig:pipeline} (top panel), conventional reward modeling approaches follow a pointwise regressive paradigm. The term "pointwise" indicates that these RMs evaluate individual images independently, while "regressive" refers to the scalar reward prediction mechanism. This paradigm encompasses both CLIP-based models, which compute rewards through cosine similarity between CLIP image and text embeddings, and VLM-based approaches, which typically employ additional regression heads to map VLM hidden states to scalar reward values. These models are commonly optimized using preference pairs with the Bradley-Terry (BT) loss:
\begin{equation}
\label{eq:bt_loss}
\mathcal{L}_{BT} = -\mathbb{E}_{(y, x^w, x^l) \sim \mathcal{D}} \left[ \log\left(\sigma\left(r(x^w, y) - r(x^l, y)\right)\right) \right],
\end{equation}
where $y$ is the input prompt, $x^w$ and $x^l$ are the chosen (better) and rejected (worse) images, $r(\cdot,\cdot)$ is the reward model, and $\sigma(\cdot)$ is the sigmoid function. These approaches ultimately provide reward signals to optimize diffusion models through reward-based feedback learning.

However, this regressive paradigm suffers from fundamental limitations. CLIP-based approaches are constrained by their dual-encoder architecture and single-modality design, which impose inherent scalability bottlenecks. For VLM-based models, the regression head introduces a critical paradigm mismatch with the model's native next-token prediction ability. Collectively, these limitations prevent full utilization of pre-trained knowledge and fundamentally restrict effective scaling across both reward  context and model size dimensions.

\subsection{Reward Model Learning}
\begin{figure}[t]
  \centering
  \includegraphics[width=1.0\textwidth, keepaspectratio]{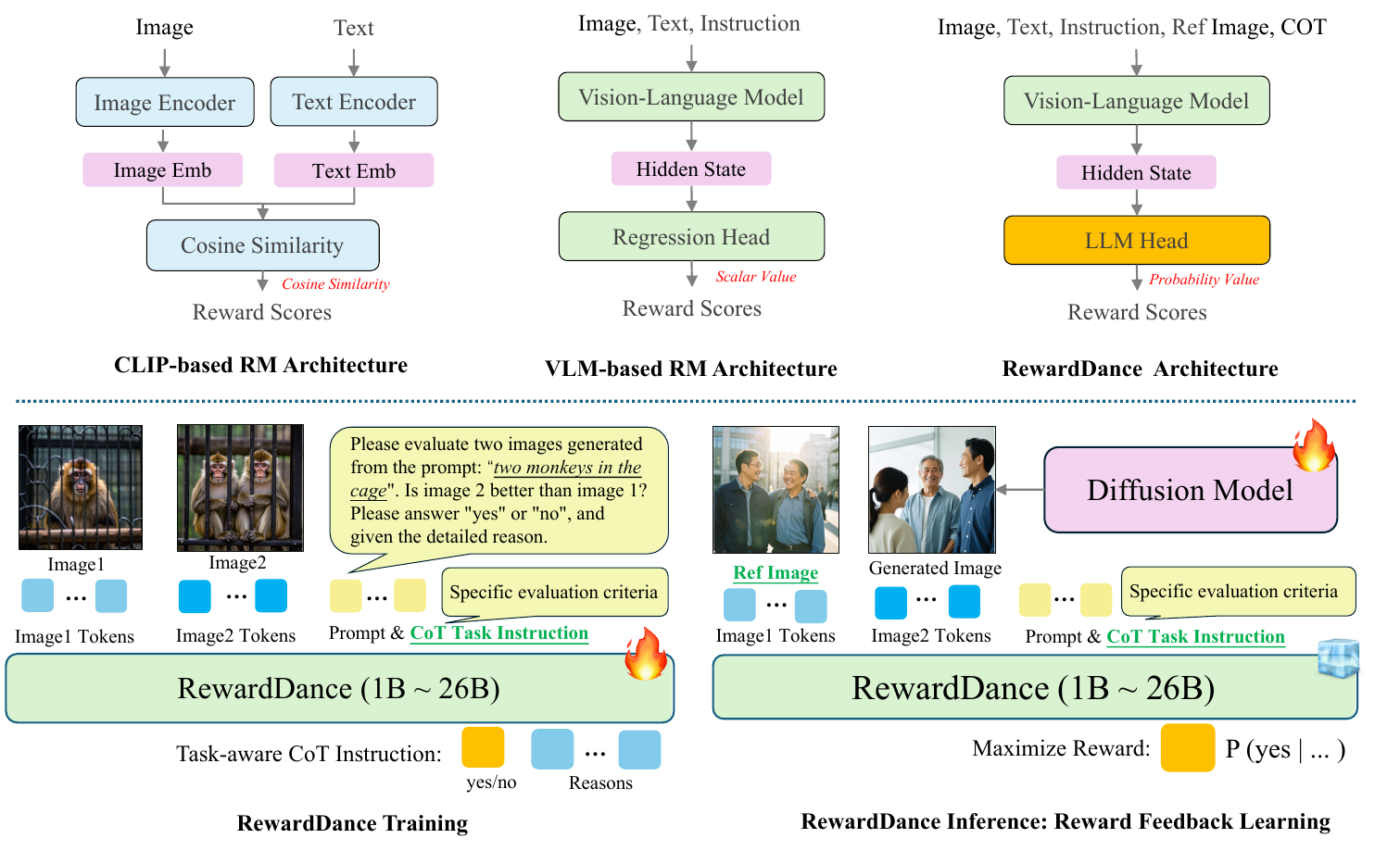}
  \vspace{-5mm}
  \caption{Overview of RewardDance framework compared with existing reward model Architecture. (Top): Previous works use CLIP-based or VLM-based reward models to provide scalar reward scores for diffusion model training. (Bottom): Our RewardDance approach (from 1B to 26B parameters) uses task-aware CoT instructions for reward modeling with reasoning. Red flames indicate trainable components; blue ice cubes indicate frozen parameters.}
  \label{fig:pipeline}
\end{figure}

To address the limitations of conventional regressive reward modeling, we introduce RewardDance, a novel framework designed for scalable visual reward modeling. Our approach centers on a generative paradigm that regards reward modeling as a token generation task, naturally aligning with the native capabilities of modern VLMs. This core innovation enables systematic exploration of scalability along two primary dimensions: Model Scaling and Context Scaling.
As shown in Figure~\ref{fig:pipeline} (bottom panel), our RewardDance framework supports models ranging from 1B to 26B parameters. The framework operates through two main stages: (1) RewardDance Training: where we train the reward model on task-aware Chain-of-Thought (CoT) instruction data, and (2) RewardDance Inference: where the trained reward model provides feedback signals to optimize diffusion models through reward feedback learning.

\subsubsection{Context Scaling}
We scale the context of our reward model beyond simple image-prompt pairs by incorporating three key elements: task-aware instructions, reference images, and Chain-of-Thought (CoT) reasoning.

First, we regard the reward scoring problem as a comparative judgment task as shown in Figure \ref{fig:pipeline} (bottom left). The model's input consists of image1 tokens $x_1$, image2 tokens $x_2$, prompt $y$, and CoT task instruction $i$. The task instruction guides the model to evaluate whether one image is superior to the other based on a set of predefined criteria. The model is then trained to predict the next token with "yes" or "no". In practice, the reward score is the predicted probability of the "yes" token:
\begin{equation}
r_{\theta}(x_1, x_2, y, i) = P_{\theta}(\text{"yes"} \mid x_1, x_2, y, i),
\label{eq:genreward}
\end{equation}
where $P_\theta$ is the probability distribution from our VLM-based reward model, RewardDance.

To further scale the context and enhance the RM's reasoning capabilities, we train it to generate a rationale for its decision in a Chain-of-Thought (CoT) manner. As illustrated in the Figure~\ref{fig:pipeline}, the model outputs both "yes/no" and "Reasons" tokens. We construct two data formats for this: one where the model outputs the "yes/no" decision before the reasoning, and another where it generates the reasoning first and concludes with the decision. This detailed reasoning data, distilled from a powerful teacher model (SEED-VL 1.5), not only improves performance but also makes the reward judgments interpretable. For efficiency during the feedback alignment stage (bottom right in the Figure~\ref{fig:pipeline}), we employ the first format and derive the reward signal by maximizing the probability of the "yes" token: $P (\text{yes} | ... )$ to optimize the diffusion model.

\begin{figure}[t]
  \centering
  \includegraphics[width=1.0\textwidth, keepaspectratio]{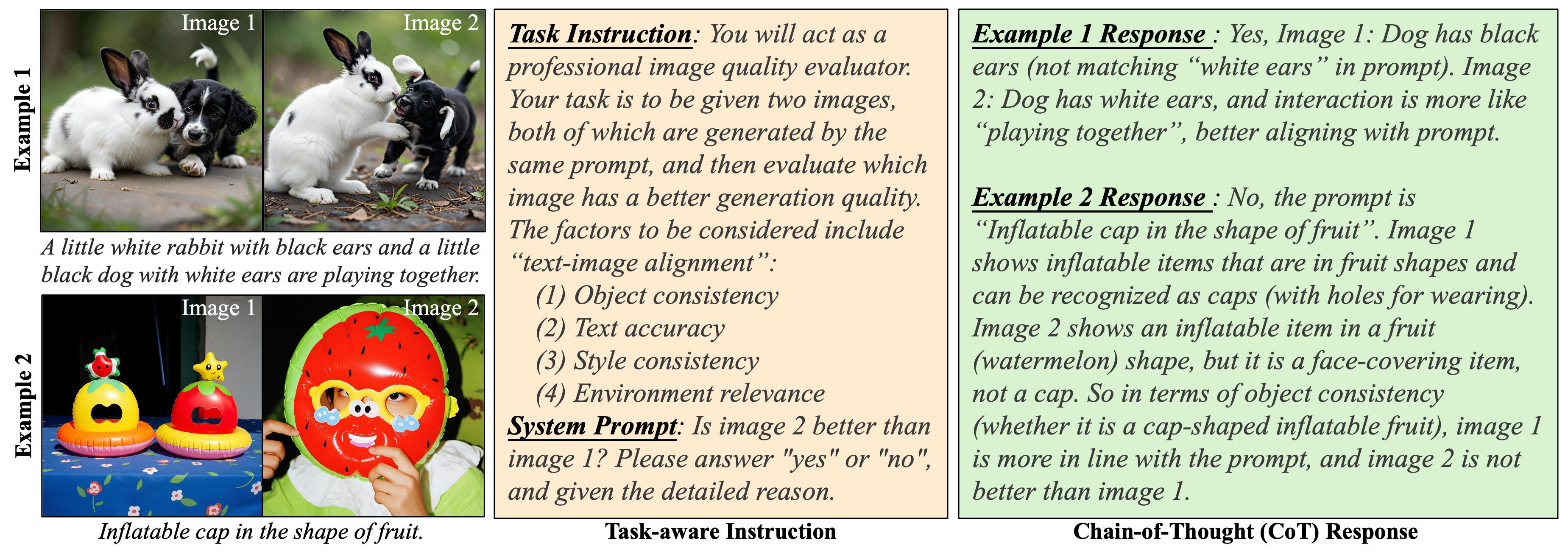}
  \vspace{-5mm}
  \caption{Examples of task-aware instruction and CoT response used for training our RewardDance model.}
  \label{fig:cot_example}
\end{figure}

Figure \ref{fig:cot_example} provides a concrete illustration of two COT examples in RM training process. It consist of a prompt , two images for comparison (Image 1 and Image 2) , a task-aware instruction, a "yes/no" judgment and a Chain-of-Thought (CoT) response to elaborate on the detailed reasoning. This structure, which combines instructions, comparative judgment, and detailed reasoning, greatly enriches the reward model's training context, allowing it to move beyond simple text-image matching to achieve more precise and interpretable evaluations.

\subsubsection{Model Scaling}
To investigate the effects of model scale, we systematically scale our reward model using variants of the InternVL~\cite{chen2024expanding} architecture, ranging from 1B to 26B parameters. Our experiments reveal a strong positive correlation between the reward model's parameter count and the final quality of generated outputs, demonstrating that larger reward models yield superior reward evaluation performance and enhanced generation quality.

\subsubsection{Reward Model Training Variants}

\textbf{Pointwise Generative Variant}:
While our primary reference (pairwise) generative reward paradigm requires reference images for reward prediction, we also design a pointwise generative variant. This variant receives only prompts and generated images, incorporating additional instructions typically used to evaluate whether generated results meet high-quality standards. The reward value is measured by the probability of the "yes" token. During training, we employ Bradley-Terry loss for preference pairs. To facilitate better model convergence, we add a weighted cross-entropy (CE) loss with a small coefficient, assigning "yes" labels to preferred samples and "no" labels to rejected samples.

\subsection{Reward Feedback Aligning}
\subsubsection{Reward Finetuning}
\label{sec:rf}
For Reinforcement Learning (RL)-based fine-tuning, we adopt the ReFL algorithm~\cite{xu2023imagereward}. Our generative reward model, RewardDance, provides crucial preference signals. Since our model operates comparatively, we utilize a Best-of-N (BoN) sampling strategy to establish high-quality references for each prompt. Specifically, we first generate N candidate images using the SFT model. RewardDance then performs pairwise comparisons among candidates to identify the top-ranked image, which serves as a reference in subsequent fine-tuning steps. We explore different strategies for selecting reference images based on win counts, with detailed experiments presented in Section~\ref{tab:exp_ab}.

\subsubsection{Inference-Time Scaling}
To further enhance inference-time performance, we implement a Search over Paths approach~\cite{liu2025inference}. This strategy works by pruning a search space of multiple generation paths to select the most promising trajectories. The process begins by initializing N distinct sampling trajectories from different noise vectors. During ODE sampling, these trajectories are iteratively extended using a re-noising and re-sampling mechanism. Specifically, a verifier (our RewardDance) prunes the search space at each step by selecting the most promising trajectories. For this verification role, we deploy a computationally efficient point-wisee generative variant of our model. This lightweight variant is optimized for speed by omitting reference images and evaluating individual candidate images based solely on task instructions.

\section{Experiments}
In this section, we present a comprehensive evaluation of RewardDance on both text-to-image, text-to-video and image-to-video tasks. We analyze the impact of reward model scaling, inference-time scaling and various architectural choices through detailed experimental results and ablation studies.

\subsection{Evaluation Baselines, Benchmarks and Metrics}

\subsubsection{\textbf{Baselines}}
\noindent{\textbf{Image Baselines.}} Our primary baselines include representative state-of-the-art (SOTA) models from academia (e.g., LlamaGen~\cite{sun2024autoregressive}, SD-v$1.5$~\cite{rombach2022high}, PixArt-$\alpha$~\cite{chen2023pixart}, SD-v$2.1$~\cite{rombach2022high}, DALL-E~$2$~\cite{ramesh2022hierarchical}, Emu$3$-Gen~\cite{wang2024emu3}, SDXL~\cite{podell2023sdxl}, DALL-E~$3$~\cite{betker2023improving}, SD-v3~\cite{esser2024scaling} and FLUX.1-dev~\cite{flux2024}) and industry (e.g., Luma~\cite{lumalabs}, Ideogram~\cite{Ideogram}, Midjourney V6.1~\cite{midjourney}, DALLE-3~\cite{betker2023improving}, FLUX1.1~\cite{flux2024}, Hunyuan~\cite{kong2024hunyuanvideo} , Imagen 3, Recraft~\cite{recraft}s, Seedream 2.0~\cite{gong2025seedream} and Qwen-Image~\cite{wu2025qwen}).

\noindent{\textbf{Video Baselines.}}
We compare our model with SOTA industry products, e.g., Kling 2.1 Master~\cite{klingai}, Veo-3.0~\cite{veo}, Kling 2.0 Master~\cite{klingai}, Veo-2.0~\cite{veo}, Wan 2.1~\cite{wan2025wan}, Sora~\cite{brooks2024video}, RunwayGen~4~\cite{runwaygen4}.

\subsubsection{\textbf{Evaluation Benchmarks}}

\noindent{\textbf{Image Evaluation Benchmarks.}}
 We adopt Bench-240, the evaluation prompts utilized in Seedream 2.0 and Seedream 3.0. This comprehensive set of 240 prompts is constructed by integrating representative prompts from publicly available benchmarks with manually curated examples. The prompt distribution is rigorously calibrated based on extensive user preference surveys. The maximum score for Bench-240 is 100 points. 

\noindent{\textbf{Video Evaluation Benchmarks.}}
For a comprehensive evaluation of video generation models across diverse scenarios, we utilize SeedVideoBench-1.0, introduced in Seedance 1.0. It comprises 300 prompts applicable to both text-to-video and image-to-video tasks, covering a wide spectrum of application domains. The taxonomy for the image-to-video task aligns with that of text-to-video, augmented with detailed annotations for the initial frame.

\subsubsection{\textbf{Evaluation Metrics}}

\noindent{\textbf{Image Metrics.}} For all text-to-image experiments, we adopt the image-text alignment score from~\cite{gao2025seedream}. This metric involves human evaluators to verify whether the generated images correctly depict the key elements described in the prompts.

\noindent{\textbf{Video Metrics.}}
 For text-to-video and image-to-video evaluation, we use the human preference-based Good-Same-Bad (GSB) score. Here, G, S, and B represent the counts of "Good," "Same," and "Bad" judgments, respectively. The final GSB score is defined as:
\begin{equation}
\text{GSB} = \frac{G-B}{G+S+B}
\label{eq:gsb}
\end{equation}

To enable performance comparison with industry products, we developed the Video-Text Alignment Score. The scoring rubric is defined as follows: 0 indicates a complete mismatch between the video content and the text description; 1 indicates a partial match; and 2 indicates a perfect match.

\subsection{Comparison with Reward Model Scaling}

\noindent{\textbf{Image Generation.}}
As demonstrated in Table~\ref{tab:image_gen}, deploying and scaling RewardDance delivers substantial performance improvements across both RL fine-tuning and test-time scaling paradigms. For RL fine-tuning, scaling the RM from 1B to 26B parameters significantly enhances FLUX.1-dev performance from 67.0 to 73.6. This scaling effect becomes even more pronounced with Seedream-3.0, where scores dramatically increase from a baseline of 74.1 to an impressive 84.8 using the 26B RM. A consistent scaling trend emerges for test-time scaling with Seedream-3.0, with performance steadily climbing from 74.1 to 80.5. This robust correlation across diverse base models and optimization paradigms confirms that larger VLM-based reward models more effectively capture human preferences, thereby guiding diffusion models toward superior output quality.

\begin{table}[htbp]
\centering
\resizebox{0.98\textwidth}{!}{%
\begin{tabular}{l|l|l|llllll}
\toprule
\textbf{Stage} & \textbf{Metric} & \textbf{Base Model} & \textbf{No RM} & \textbf{1B RM} & \textbf{2B RM} & \textbf{4B RM} & \textbf{8B RM} & \textbf{26B RM} \\
\midrule
RM Training & RM ID Accuracy (\%) & RewardDance & -- & 64.70 & 69.36 & 65.37 & 74.92 & 78.44 \\
RM Training & RM OOD Accuracy (\%) & RewardDance & -- & 69.10 & 69.59 & 71.92 & 71.94 & 80.90 \\
\toprule
\multirow{2}{*}{RL Fine-tuning} & Alignment Score & FLUX.1-dev & 67.0 & $70.7_{\textcolor{ForestGreen}{+3.7}}$ & $72.4_{\textcolor{ForestGreen}{+5.4}}$ & $72.2_{\textcolor{ForestGreen}{+5.2}}$ & $73.0_{\textcolor{ForestGreen}{+6.0}}$ & $73.6_{\textcolor{ForestGreen}{+6.6}}$ \\
& Alignment Score & Seedream-3.0-SFT & 74.1 & $74.9_{\textcolor{ForestGreen}{+0.8}}$ & $75.3_{\textcolor{ForestGreen}{+1.2}}$ & $79.5_{\textcolor{ForestGreen}{+5.4}}$ & $81.6_{\textcolor{ForestGreen}{+7.5}}$ & $84.8_{\textcolor{ForestGreen}{+10.7}}$ \\
\midrule
Test-Time Scaling & Alignment Score & Seedream-3.0-SFT & 74.1 & $75.1_{\textcolor{ForestGreen}{+1.0}}$ & $76.3_{\textcolor{ForestGreen}{+2.2}}$ & $78.4_{\textcolor{ForestGreen}{+4.3}}$ & $79.3_{\textcolor{ForestGreen}{+5.2}}$ & $80.5_{\textcolor{ForestGreen}{+6.4}}$ \\
\bottomrule
\end{tabular}%
}
\caption{The impact of Reward Model (RM) scaling on both RM accuracy and final text-to-image alignment score. As the RM size and its corresponding accuracy increase, the performance of the diffusion models under both RL and Test-time Scaling consistently improves.}
\label{tab:image_gen}
\end{table}

Furthermore, we constructed two evaluation datasets: In-Domain (ID) preference dataset contains 2,500 sample pairs. This dataset was constructed by partitioning a subset of data points from the Stage 2 training set of RewardDance that were held out and unseen during training. Out-Of-Domain (OOD) preference dataset contains over 4,000 sample pairs. This dataset was curated by sampling from public benchmark datasets, including ImageReward~\cite{xu2023imagereward} and HPS~\cite{wu2023better}, to assess the model's generalization capability on out-of-distribution samples.

As shown in Table~\ref{tab:image_gen}, we observe no strict positive correlation between the Reward Model (RM) accuracy on the ID dataset and its parameter scale, particularly evident in models with 1B, 2B, and 4B parameters. This finding aligns with existing studies~\cite{chen2024accuracy, razin2025makes, wen2024rethinking}, which suggest that higher RM accuracy does not guarantee corresponding improvements in Reinforcement Learning (RL) performance.
However, the RM's accuracy on the OOD dataset – indicative of its generalization capability – can be viewed  as a more significant predictor of the final outcome. An RM that demonstrates accurate judgment on unseen distributions and strong generalization capability possesses substantially higher guiding value. Consequently, for evaluating RM efficacy, OOD accuracy (generalization capability) emerges as a more critical core metric than ID accuracy.
Building upon this insight, developing evaluation benchmarks that better assess RM generalization constitutes an important direction for future research. This involves enhancing the consistency between the RM evaluation set and the RL test set, as well as integrating sample pairs of varying difficulty levels to enable a more comprehensive assessment of reward model capabilities.

\noindent{\textbf{Video Generation.}}
We demonstrate that the benefits of reward scaling extend seamlessly to video generation tasks, as presented in Table~\ref{tab:video_gen}. We employ GSB metric for this evaluation. For Seedance-1.0 in the Text-to-Video (T2V) RL setting, RewardDance exhibits a compelling scaling trend: performance improves by +28\% with a 1B RM compared to Seedance-1.0-SFT,  and progressively increases to an impressive +49\% improvement when utilizing the 26B RM. This scaling phenomenon is equally pronounced in the Image-to-Video (I2V) paradigm, where performance gains systematically increase from +29\% with the 1B RM to a substantial +47\% with the 26B RM. These consistent improvements across both video generation modalities demonstrate the broad applicability and robustness of our RewardDance framework.

\begin{table}[htbp]
\centering
\resizebox{0.98\textwidth}{!}{%
\begin{tabular}{c|c|clllll}
\toprule
\textbf{Task} & \textbf{Base Model} & \textbf{No RM} & \textbf{1B RM} & \textbf{2B RM} & \textbf{4B RM} & \textbf{8B RM} & \textbf{26B RM} \\
\midrule
T2V RL & Seedance-1.0-SFT & - & +28\% &  $+32\%_{\textcolor{ForestGreen}{+4\%}}$ & $+41\%_{\textcolor{ForestGreen}{+13\%}}$ & $+45\%_{\textcolor{ForestGreen}{+17\%}}$ & $+49\%_{\textcolor{ForestGreen}{+21\%}}$ \\
I2V RL & Seedance-1.0-SFT & - & +29\% & $+34\%_{\textcolor{ForestGreen}{+5\%}}$ & $+37\%_{\textcolor{ForestGreen}{+8\%}}$ & $+41\%_{\textcolor{ForestGreen}{+12\%}}$ & $+47\%_{\textcolor{ForestGreen}{+18\%}}$ \\
\bottomrule
\end{tabular}
}
\caption{Video text alignment results (using GSB as the metric), showing that increasing the reward model size yields significant and consistent performance gains for both T2V and I2V RL over the SFT baseline.}
\label{tab:video_gen}
\end{table}

\subsection{Comparison with State-of-the-art Generation Models}
To rigorously assess the competitiveness of our framework, we benchmark our RewardDance-optimized models against leading academic and industrial systems on three benchmarks: GenEval, Bench-240, and SeedVideoBench-1.0. As shown in Tables~\ref{tab:geneval}, \ref{tab:comparision_with_sota_image}, and \ref{tab:comparision_with_sota_video}, models enhanced by RewardDance consistently achieve state-of-the-art performance.

\noindent{\textbf{GenEval.}}
As shown in Table~\ref{tab:geneval}, models enhanced with RewardDance demonstrate significant superiority across multiple dimensions of text-to-image generation. Seedream-3.0 w RewardDance achieves the top overall score of 0.79, while FLUX.1-dev w RewardDance follows closely at 0.75, outperforming strong baselines like SD3 (0.74). RewardDance confers overall performance gains of +0.10 and +0.09 on Seedream-3.0 w/o RewardDance and FLUX.1-dev, respectively. These advancements validate the efficacy of RewardDance in substantially enhancing semantic understanding and generative precision.

\begin{table}[!h]
    \centering
    \setlength{\tabcolsep}{3mm}
    \resizebox{0.98\linewidth}{!}{
    \begin{tabular}{llllllll}
        \toprule
        \textbf{Method}  & \textbf{Single Obj.} & \textbf{Two Obj.} & \textbf{Counting} & \textbf{Colors} & \textbf{Position} & \textbf{Color Attri.} & \textbf{Overall$\uparrow$} \\
        \midrule
        LlamaGen~\cite{sun2024autoregressive}  & $0.71$ & $0.34$ & $0.21$ & $0.58$ & $0.07$ & $0.04$ & $0.32$ \\
        SD-v$1.5$~\cite{rombach2022high} &  $0.97$ & $0.38$ & $0.35$ & $0.76$ & $0.04$ & $0.06$ & $0.43$ \\
        PixArt-$\alpha$~\cite{chen2023pixart} &  $0.98$ & $0.50$ & $0.44$ & $0.80$ & $0.08$ & $0.07$ & $0.48$ \\
        SD-v$2.1$~\cite{rombach2022high} &  $0.98$ & $0.51$ & $0.44$ & $0.85$ & $0.07$ & $0.17$ & $0.50$ \\
        DALL-E~$2$~\cite{ramesh2022hierarchical}  & $0.94$ & $0.66$ & $0.49$ & $0.77$ & $0.10$ & $0.19$ & $0.52$ \\
        Emu$3$-Gen~\cite{wang2024emu3}  & $0.98$ & $0.71$ & $0.34$ & $0.81$ & $0.17$ & $0.21$ & $0.54$ \\
        SDXL~\cite{podell2023sdxl} &  $0.98$ & $0.74$ & $0.39$ & $0.85$ & $0.15$ & $0.23$ & $0.55$ \\
        DALL-E~$3$~\cite{betker2023improving}  & $0.96$ & $0.87$ & $0.47$ & $0.83$ & $0.43$ & $0.45$ & $0.67$ \\
        SD-v3~\cite{esser2024scaling} & 0.99 & 0.94 & 0.72 & 0.89 & 0.33 & 0.60 & $0.74$ \\
        \midrule
        FLUX.1-dev~\cite{flux2024} & 0.98 & 0.81 & 0.74 & 0.79 & 0.22 & 0.45 & 0.66 \\
        \textbf{FLUX.1-dev w RewardDance } & $\textbf{0.98}_{\textcolor{ForestGreen}{+0.00}}$ & $\textbf{0.92}_{\textcolor{ForestGreen}{+0.11}}$ & $\textbf{0.83}_{\textcolor{ForestGreen}{+0.09}}$ & $\textbf{0.84}_{\textcolor{ForestGreen}{+0.05}}$ & $\textbf{0.30}_{\textcolor{ForestGreen}{+0.08}}$ & $\textbf{0.64}_{\textcolor{ForestGreen}{+0.19}}$ & $\textbf{0.75}_{\textcolor{ForestGreen}{+0.09}}$ \\
        \midrule
        Seedream-3.0~\cite{gao2025seedream} SFT (w/o RewardDance) & 0.98 & 0.86 & 0.67 & 0.86 & 0.43 & 0.31 & 0.69 \\
        \textbf{Seedream-3.0 w RewardDance} & $\textbf{1.00}_{\textcolor{ForestGreen}{+0.02}}$ & $\textbf{0.96}_{\textcolor{ForestGreen}{+0.10}}$ & $\textbf{0.88}_{\textcolor{ForestGreen}{+0.21}}$ & $\textbf{0.94}_{\textcolor{ForestGreen}{+0.08}}$ & $\textbf{0.52}_{\textcolor{ForestGreen}{+0.09}}$ & $\textbf{0.44}_{\textcolor{ForestGreen}{+0.13}}$ & $\textbf{0.79}_{\textcolor{ForestGreen}{+0.10}}$ \\
        \bottomrule
    \end{tabular}
    }
    \caption{Evaluation of text-to-image generation ability on GenEval benchmark.}
    \label{tab:geneval}
\end{table}

\noindent{\textbf{Bench-240.}}
On the Bench-240 benchmark (Table~\ref{tab:comparision_with_sota_image}), Seedream-3.0 w RewardDance achieves the highest Overall Score of 0.848 when compared against a suite of powerful, closed-source models. It surpasses top commercial offerings such as Imagen 3 (0.79), Luma (0.77), and Midjourney V6.1 (0.63). A detailed breakdown reveals that our model demonstrates exceptional performance in complex categories like Action (0.87) and is highly competitive in Attribute understanding (0.89), validating the effectiveness of our training and alignment methodology.

\begin{table}[!ht]
    \centering
    \begin{tabular}{lccccc}
        \toprule
        \textbf{Model} & \textbf{Overall Score} & \textbf{Action} & \textbf{Counting} & \textbf{Relation} & \textbf{Attribute} \\
        \midrule
        Hunyuan~\cite{kong2024hunyuanvideo} & 0.59 & 0.59 & 0.60 & 0.48 & 0.65 \\
        Midjourney V6.1~\cite{midjourney} & 0.63 & 0.62 & 0.51 & 0.52 & 0.68 \\
        Recraft~\cite{recraft} & 0.64 & 0.58 & 0.50 & 0.51 & 0.68 \\
        DALLE-3~\cite{betker2023improving} & 0.67 & 0.70 & 0.57 & 0.53 & 0.71 \\
        FLUX1.1~\cite{flux2024} & 0.67 & 0.68 & 0.56 & 0.62 & 0.74 \\
        Seedream 2.0~\cite{gong2025seedream} & 0.71 & 0.67 & 0.67 & 0.66 & 0.77 \\
        Qwen-Image~\cite{wu2025qwen} & 0.77 & 0.82 & 0.73 & 0.76 & 0.88 \\
        Luma~\cite{lumalabs} & 0.77 & 0.71 & 0.68 & \textbf{0.81} & 0.82 \\
        Ideogram~\cite{Ideogram} & 0.77 & 0.78 & 0.70 & 0.69 & 0.80 \\
        Imagen 3 & 0.79 & 0.77 & 0.66 & 0.73 & \textbf{0.90} \\
        \midrule 
        \textbf{Seedream3.0  ~\cite{gao2025seedream} w RewardDance}  & \textbf{0.848} & \textbf{0.87} & \textbf{0.78} & \textbf{0.81} & 0.89 \\
        \bottomrule
    \end{tabular}
    \caption{Comparison of human evaluation of different generation models on text-to-image generation task.}
\label{tab:comparision_with_sota_image}
\end{table}

\noindent{\textbf{SeedVideoBench-1.0.}}
We also benchmark our method with Video-Text Alignment Score on video generation tasks (Table~\ref{tab:comparision_with_sota_video}), evaluating our RewardDance-aligned Seedance 1.0 against leading proprietary models. 
The final score is the average rating calculated over all samples. In the Text-to-Video (T2V) evaluation, Seedance 1.0 achieves the highest average score of 1.66, outperforming strong competitors like Veo-3.0 (1.63) and Kling 2.1 (1.57). For the Image-to-Video (I2V) task, our model is also state-of-the-art, achieving a top-tier score of 1.65 and tying with the best-performing model, Kling 2.1. These findings underscore our framework's ability to produce highly competitive, state-of-the-art video generations.

\begin{table}[!ht]
    \centering
    \begin{tabular}{l cc}
        \toprule
        \textbf{Model} & \textbf{T2V Avg. Score} & \textbf{I2V Avg. Score} \\
        \midrule
        RunwayGen 4~\cite{runwaygen4} & -- &1.37 \\
        Sora~\cite{brooks2024video} & 1.37 & -- \\
        Veo-2.0~\cite{veo} & 1.47 & 1.19 \\
        Wan 2.1~\cite{wan2025wan} & 1.49 & 1.36 \\
        Kling 2.1 Master~\cite{klingai} & 1.57 & 1.65 \\
        Kling 2.0 Master~\cite{klingai} & 1.58 & 1.57 \\
        Veo-3.0~\cite{veo} & 1.63 & 1.59 \\
        \midrule 
        \textbf{Seedance 1.0 ~\cite{gao2025seedance} w RewardDance} & \textbf{1.66} & \textbf{1.65} \\
        \bottomrule
    \end{tabular}
    \caption{Comparison of human evaluation of different models on text-to-video and image-to-video generation tasks.}
    \label{tab:comparision_with_sota_video}
\end{table}

\subsection{Ablation Study}
\label{tab:exp_ab}

\begin{figure}[hp]
  \centering
  \includegraphics[width=1.0\textwidth, keepaspectratio]{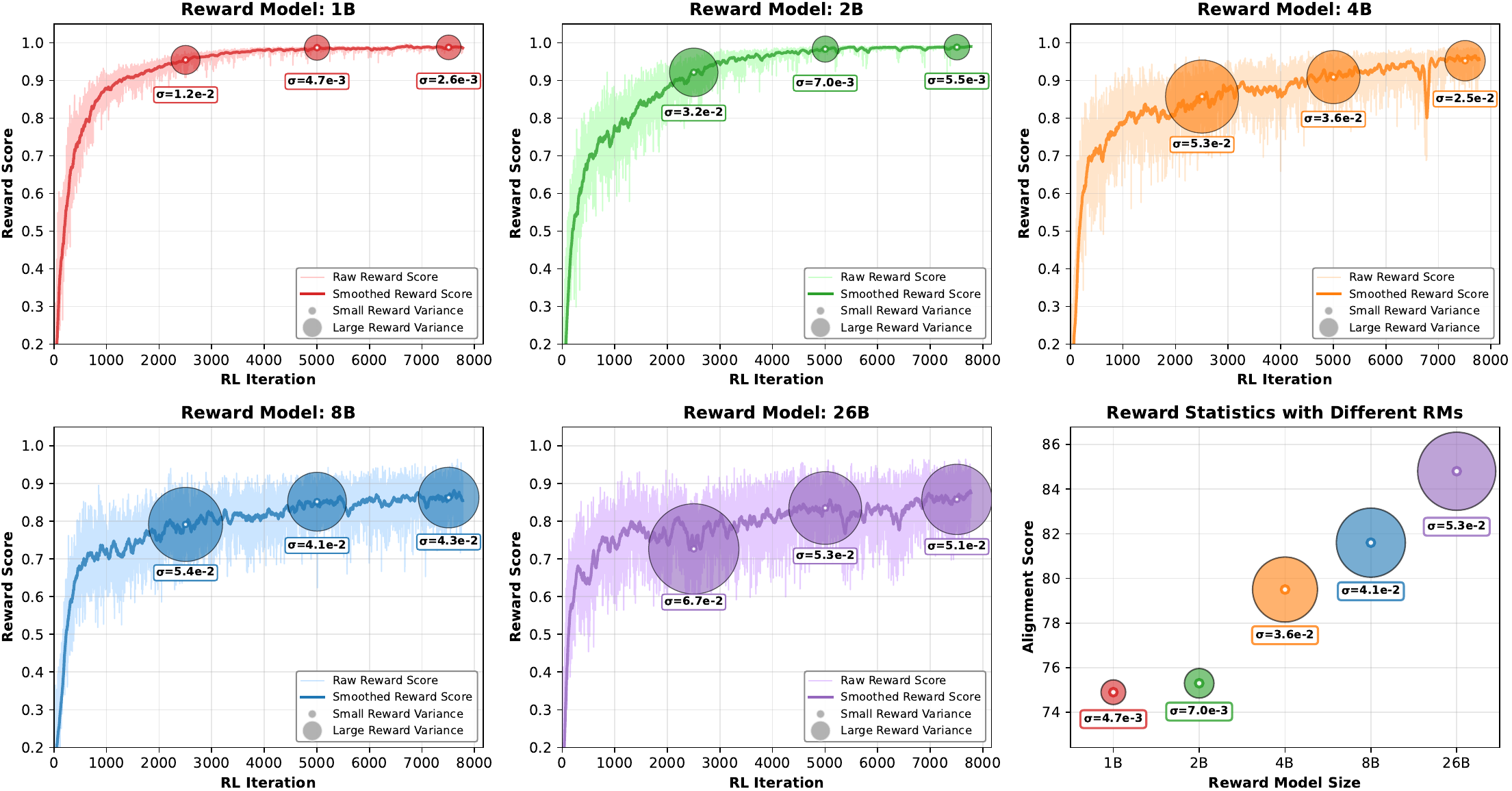}
  \vspace{-5mm}
  \caption{This figure shows the reward curves for Seedream during the RL stage, experiments with a separate reward model of varying sizes (1B to 26B).  While reward scores consistently improve with more RL iterations across all models, a key trade-off emerges with RM size: larger RMs tend to exhibit a higher standard deviation, suggesting stronger robustness and less susceptibility to reward hacking.}
  \label{fig:seedream_reward_statistics}
\end{figure}

\begin{figure}[tbp]
  \centering
  \includegraphics[width=1.0\textwidth, keepaspectratio]{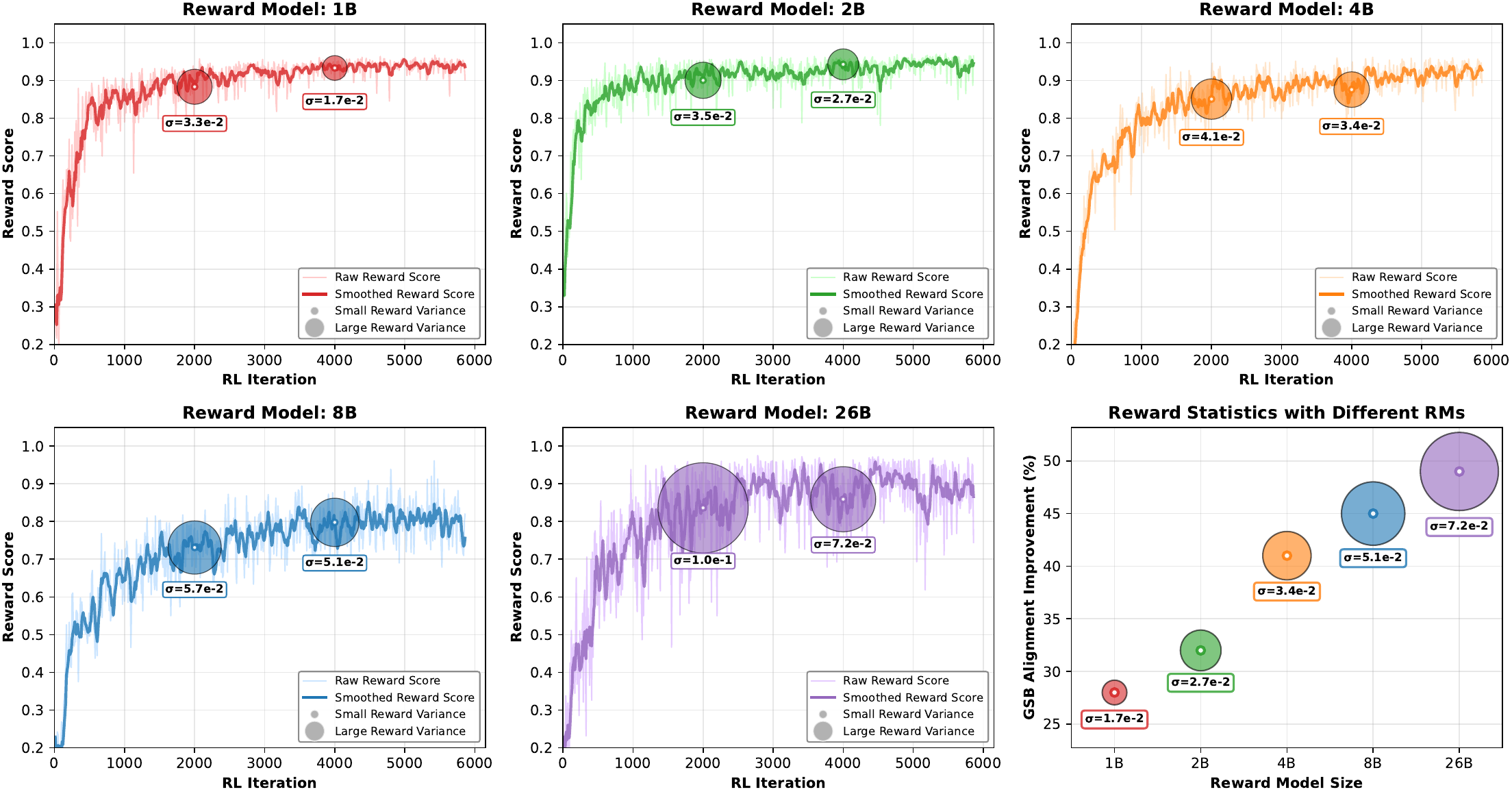}
  \vspace{-5mm}
  \caption{This figure shows the GSB alignment scores for Seedance during the RL stage, experiments with a separate reward model of varying sizes (1B to 26B). Note that the apparent sparsity of fluctuations in this plot (compared to the image task results in Figure~\ref{fig:seedream_reward_statistics}) stems from the lower sampling frequency of data points recorded for the video task. While reward scores consistently improve with more RL iterations, which phenomenon is consistent with Seedream.}
  \label{fig:seedance_reward_statistics}
\end{figure}

\textbf{Reward Dynamics Analysis.} 
Figure~\ref{fig:seedream_reward_statistics} and Figure~\ref{fig:seedance_reward_statistics}  (first five line graphs) illustrates the reward curves during ReFL training for reward models of varying scales across image and video tasks. In these visualizations, solid lines represent smoothed reward trajectories, and the surrounding light-colored bands capture the actual fluctuation ranges of raw reward values. The bubbles quantify reward variance magnitude within a sliding window of 1,000 RL iteration, simultaneously reflecting the exploration intensity of the current policy.

Key findings from these analyses include: 
i) \textit{Sustained Exploration in Large-Scale Reward Models}: The 26B RM exhibits substantially greater variance throughout the training. Critically, even at advanced training iterations, these error bars remain considerably wide, indicating that the model maintains significant exploration capabilities well into the later training phases. ii) \textit{Accelerated Convergence in Smaller Models}: In contrast, error bars for 2B/1B RMs converge to narrow ranges much earlier in training (virtually disappearing in later stages), signifying that their policies achieve stable convergence states sooner and cease effective exploration. iii) \textit{Positive Scaling-Exploration Correlation}: Consistently, larger model scales correspond to wider error bars across all experimental conditions. This strongly suggests that large-scale reward models demonstrate greater resistance to reward hacking and maintain superior sustained exploration capacity.

\begin{figure}[tbp]
\centering
\begin{minipage}[t]{0.5\textwidth}
    \vspace{0pt} 
    \centering
    \resizebox{\linewidth}{!}{%
        \begin{tabular}{c|c|c|c}
        \toprule
        \textbf{Base Model} & \textbf{Paradigm} & \makecell{\textbf{Reference}\\\textbf{Examples}} & \makecell{\textbf{Image-Text}\\\textbf{Alignment}} \\
        \midrule
        \multirow{3}{*}{FLUX.1-dev~\cite{flux2024}} &Pointwise Regressive & $\times$ & 70.8 \\
        &Pointwise Generative & $\times$  & 71.6 \\
        &Pairwise Generative & \checkmark  & \textbf{73.0} \\
        \midrule
        \multirow{3}{*}{Seedream-3.0 SFT} &Pointwise Regressive & $\times$  & 80.7 \\
        &Pointwise Generative & $\times$ & 81.0 \\
        &Pairwise Generative & \checkmark  & \textbf{81.6} \\
        \bottomrule
        \end{tabular}%
    }
      \vspace{-3mm}
    \captionof{table}{Both generative reward modeling paradigm and reward context scaling consistently improve performance.}
    \label{tab:ablation_summary}
    \vspace{1em}
    \resizebox{\linewidth}{!}{%
        \begin{tabular}{c|c|c}
        \toprule
        \scriptsize
        \textbf{Base Model} & \textbf{Type} & \makecell{\textbf{Image-Text}\\\textbf{Alignment}} \\
        \midrule
        \multirow{4}{*}{Seedream-3.0 SFT} & Best-of-16 Top-2 & 83.6 \\
         & Best-of-16 Bottom-2  & 80.6 \\
         & Best-of-2 Top-2 & 82.7 \\
         & Best-of-6 Top-2 & 83.1 \\
        \bottomrule
        \end{tabular}%
    }
          \vspace{-3mm}
    \captionof{table}{Ablation of the BON Reference.}
    \label{tab:bon}
    \vspace{1em}
    \resizebox{\linewidth}{!}{%
        \begin{tabular}{c|c|c}
        \toprule
        \textbf{Base Model} & \textbf{Type} & \textbf{Image-Text Alignment} \\
        \midrule
        \multirow{3}{*}{Seedream-3.0 SFT} & Baseline & 81.6 \\
         & +CoT Finetuing  & 83.6 \\
        \bottomrule
        \end{tabular}
    }
          \vspace{-3mm}
    \captionof{table}{Ablation of COT Finetuning.}
    \label{tab:cot_domain}
\end{minipage}%
\hfill
\begin{minipage}[t]{0.48\textwidth}
    \vspace{0pt} 
    \centering
    \includegraphics[width=\linewidth, keepaspectratio]{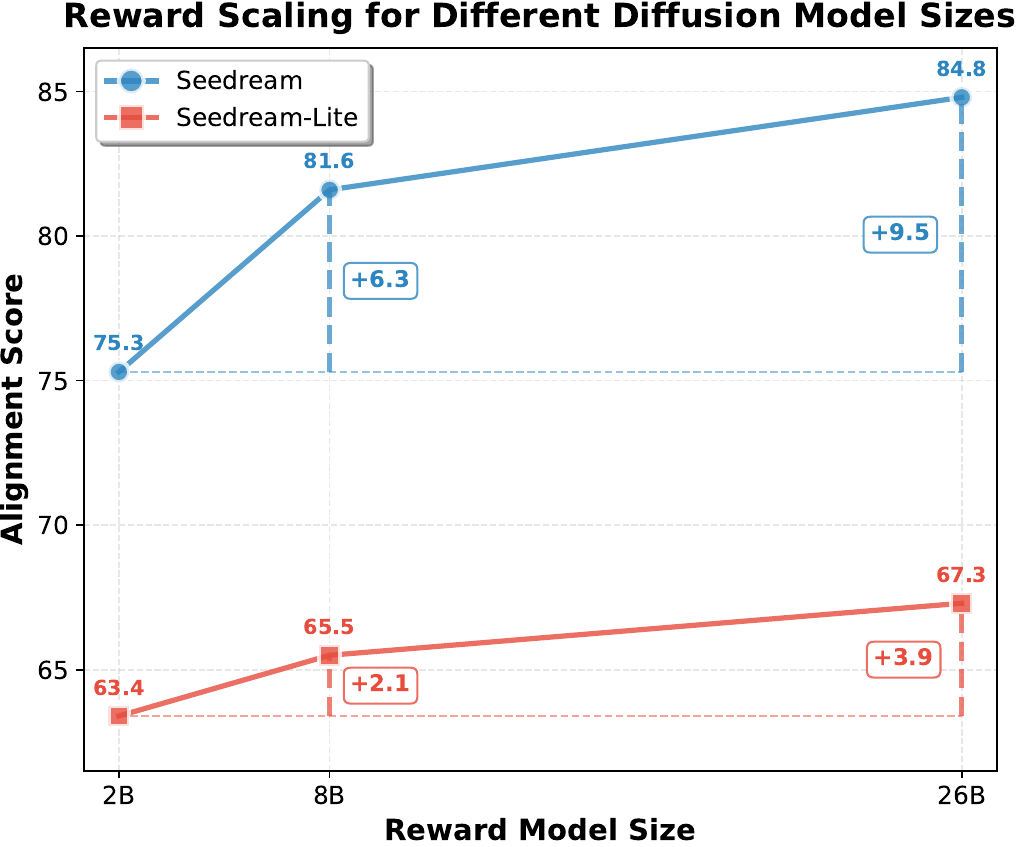}
    \captionof{figure}{The performance of Seedream and Seedream-Lite models with different RM model sizes. While both models benefit from larger RMs, the performance improvement is substantially more pronounced for the larger model. This suggests that larger generative models require commensurately larger reward models to realize their full potential.}
    \label{fig:seedream_small-large_comparison}
\end{minipage}
\end{figure}

Figure~\ref{fig:seedream_reward_statistics} and Figure~\ref{fig:seedance_reward_statistics} (final bubble chart) further visualizes the relationship between final performance metrics and exploration intensity (measured by reward variance in late training stages) across RMs of varying scales. This analysis corroborates our observations: as RM scale increases, both model performance and exploration capability (variance) improve monotonically. These concurrent improvements across both dimensions highlight the advantages of large-scale RMs in preserving exploration vitality while achieving superior final performance.

\textbf{DIT Parameter Analysis.} 
Figure~\ref{fig:seedream_small-large_comparison} presents performance curves for Seedream-3.0 Lite (approximately one-fifth the parameters of Seedream-3.0) and Seedream-3.0 under reward models of varying scales. Key observations include: i)  \textit{Scaling Laws Apply to Smaller Architectures}: Compact DiT models (Seedream-3.0 Lite) also demonstrate scaling phenomena, where performance improves with increasing RM size, consistent with their larger counterparts. ii)  \textit{Enhanced Scaling Benefits for Larger Models}: However, the performance gain rate for Seedream-3.0 Lite does not exceed that of the full-scale model. Specifically, when scaling the RM from 8B to 26B parameters: Seedream-3.0 achieves a performance improvement of +3.92\% (81.6\% → 84.8\%), while Seedream-3.0 Lite exhibits a more modest gain of +2.75\% (65.5\% → 67.3\%). This comparative analysis demonstrates that larger DiT architectures derive greater benefits from reward scaling, yielding more substantial performance improvements.

\textbf{Generative Reward Paradigm.}
We first evaluate the impact of paradigm shift from regression-based to generation-based reward modeling via our 8B RewardDance. As demonstrated in Table~\ref{tab:ablation_summary}, transitioning from regressive to generative paradigms consistently improves performance across both FLUX.1-dev (70.8 → 71.6) and Seedream (80.7 → 81.0). Subsequently, incorporating reference images yields further improvements: FLUX.1-dev advances from 71.6 to 73.0, while Seedream progresses from 81.0 to 81.6. These results validate our hypothesis that generative approaches, which align more naturally with VLM architectures, provide a more effective foundation for reward modeling and scaling.

\textbf{BON Reference Examples.}
Building upon the generative paradigm, we assess the effectiveness of incorporating Best-of-N (BoN) reference examples for enabling comparative evaluations. Due to computational constraints in image generation, we limit N to a maximum of 16. Table~\ref{tab:bon} reveals that higher-quality reference images correspond to incremental performance enhancements in Seedream-3.0 (Best-of-16 Top-2 > Best-of-6 Top-2 > Best-of-2 Top-2 > Best-of-16 Bottom-2), where 'Best-of-16 Top-2' refers to selecting the two best-quality samples from 16 inferred samples using the method described in \ref{sec:rf}. This indicates that the quality of the reference images is of critical importance in comparative reward modeling.

\textbf{Reasoning.}
We isolate the benefits of training with rich, explanatory reasoning data. As shown in Table \ref{tab:cot_domain}, incorporating CoT data provides substantial performance improvements, elevating the Image-Text Alignment score from 81.6 to 83.6 for Seedream-3.0. This demonstrates that leveraging explicit reasoning pathways enhances the reward model's capability to generate more accurate and human-aligned judgments.

\subsection{Visualizations}

\begin{figure}[t]
  \centering
  \includegraphics[width=1.0\textwidth, keepaspectratio]{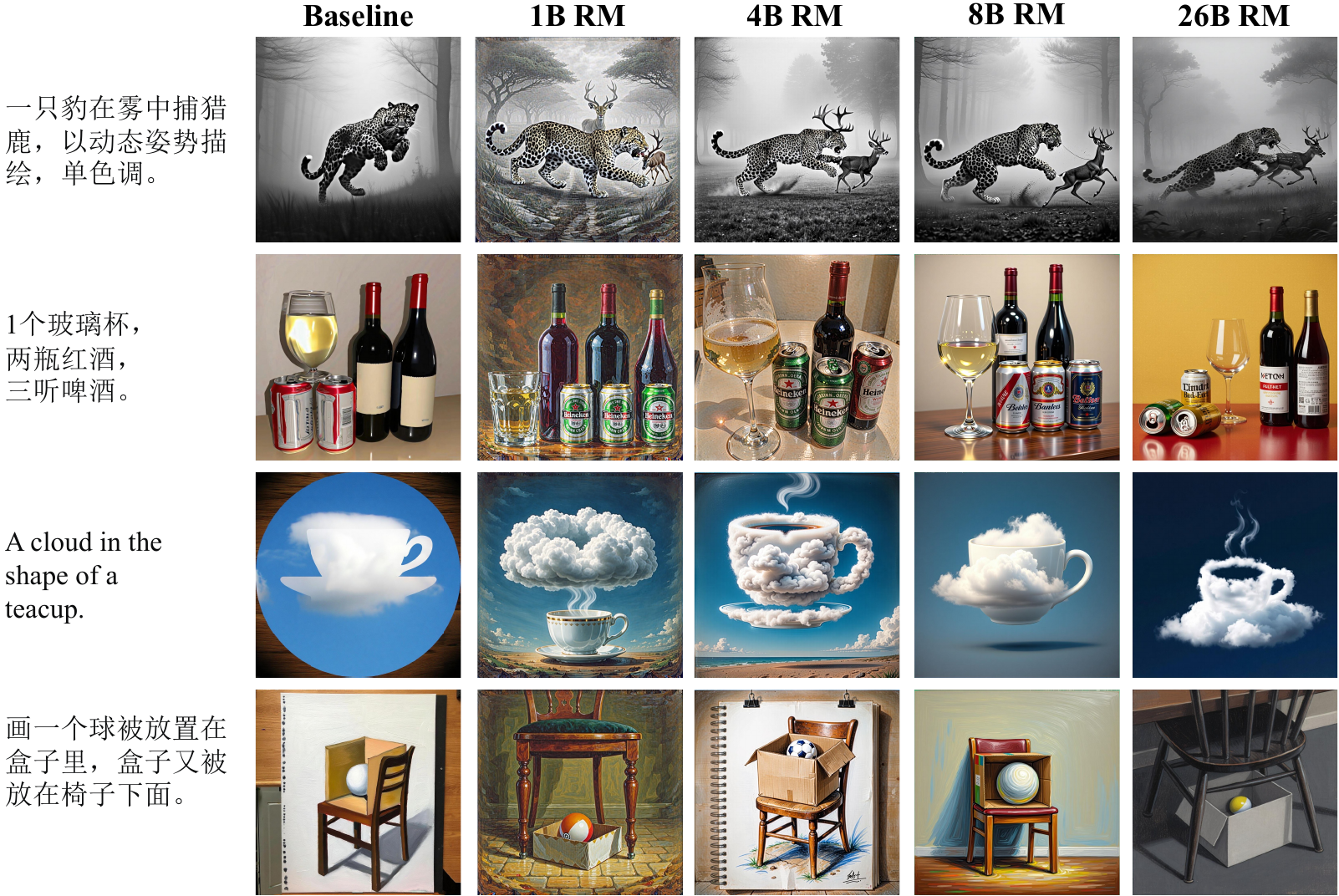}
  \vspace{-5mm}
  \caption{Text-to-Image generation comparison across reward models of increasing size (Baseline, 1B, 4B, 8B, 26B). Larger reward models demonstrate progressively better prompt adherence, visual quality, and semantic understanding.}
  \label{fig:image_rm_scaling_comparison}
\end{figure}

\begin{figure}[htbp]
  \centering
  \includegraphics[width=1.0\textwidth, keepaspectratio]{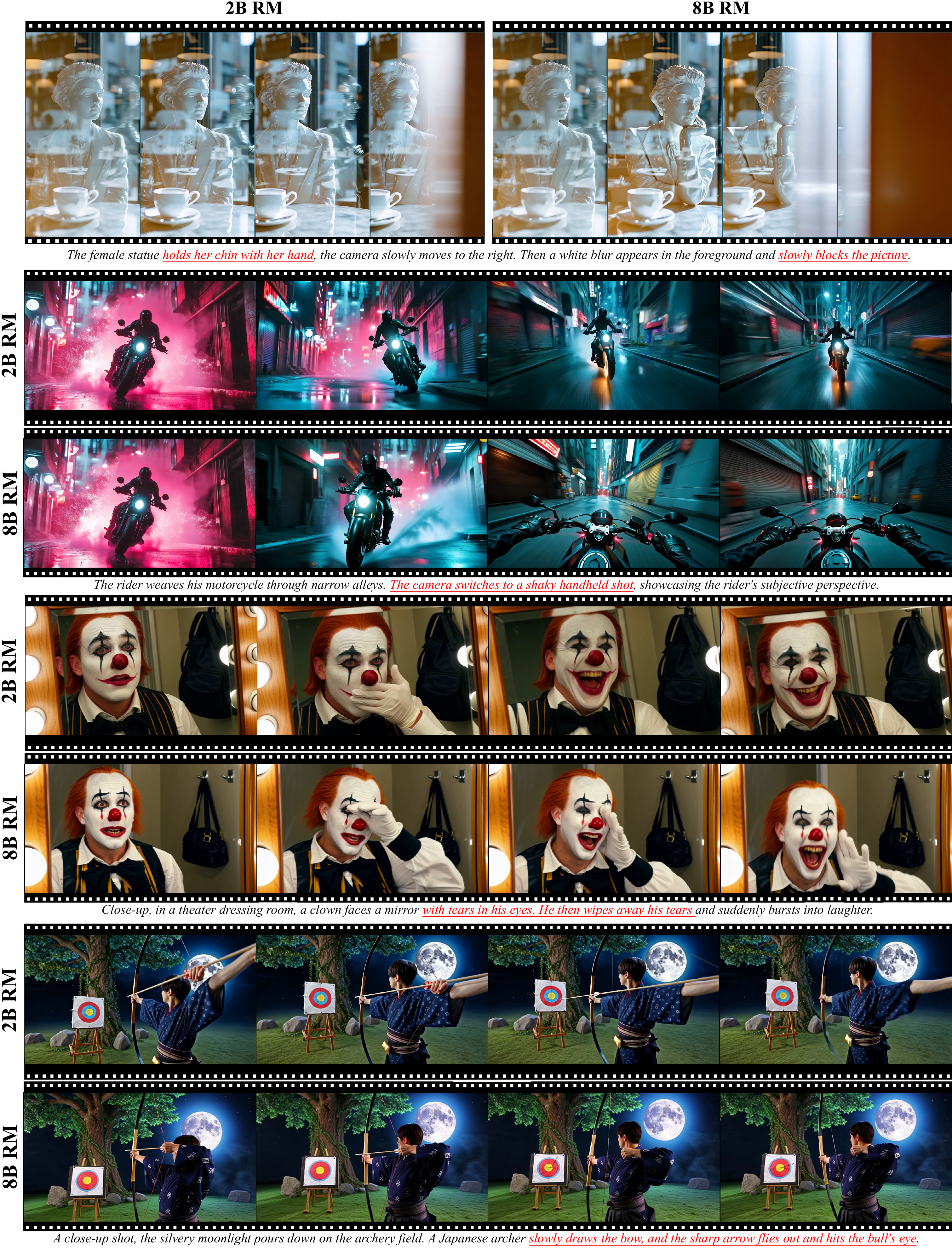}
  \vspace{-5mm}
  \caption{Video generation quality comparison between 2B and 8B reward models. Top two examples: Image-to-Video generation; Bottom two examples: Text-to-Video generation. The 8B reward model shows improved visual quality and temporal consistency compared to the 2B reward model.}
  \label{fig:video_rm_scaling_comparison}
\end{figure}

Figure \ref{fig:image_rm_scaling_comparison} and Figure \ref{fig:video_rm_scaling_comparison} visualize the effects of Reward Scaling on the image and video tasks, respectively. It can be observed from the figures that as the RM model size increases, the performance on both image and video tasks shows an improving trend. Taking the image task as an example, smaller-scale models struggle to accurately generate descriptions involving multi-instance quantity relationships, whereas the large 26B model is capable of generating them completely and correctly.

\section{Discussion and Further work}

Although RewardDance has achieved significant progress on image and video generation tasks, substantial room for exploration remains:

\textbf{Parameter Scaling.} The current maximum model size is 26B parameters. Empirical observations suggest that further scaling to larger sizes (e.g., 70B/100B) is likely to yield greater performance gains, which we will investigate in future work.

\textbf{Capability Dimension Scaling.} This work focuses mainly on foundational vision language capabilities, such as alignment. Future research will explore other critical dimensions of visual generation, such as motion modeling and aesthetic generation.

\textbf{Task Scope Scaling.} Unified models for joint understanding and generation are a current research hotspot. An important direction is to explore whether Reward Scaling techniques can be effectively applied to such models to synergistically enhance performance across tasks like visual understanding, generation, and editing.

\textbf{Multimodal Scaling.} Future many-to-vision tasks (e.g., audio/video-to-video generation) are expected to become mainstream. Consequently, advancing Cross-modal Reward Signal Scaling is a crucial research trend.

\textbf{Context Scaling.} Integrating richer reference information, more complex instructions, reflection capabilities, and in-context learning mechanisms holds promise for further boosting model performance.

\section{Conclusion}
In this work, we address a critical gap in visual Reward Models (RMs) for diffusion-based generation, where existing methods are constrained by architectural limitations or paradigm mismatches that preclude effective reward scaling. We introduce RewardDance, a scalable RM framework built on a novel generative paradigm. This approach reframes reward prediction as a token generation task by converting reward scores into a Vision-Language Model's (VLM's) predicted probability for a "yes" token, which natively aligns with the VLM's autoregressive mechanism. This core innovation unlocks scaling along two key dimensions: model size (from 1B to 26B parameters) and context richness (task-aware instructions, reference examples, and Chain-of-Thought reasoning). Our comprehensive experiments across text-to-image, text-to-video, and image-to-video tasks demonstrate that scaling along these dimensions consistently improves the quality of the reward signal, which in turn drives stable and significant gains in final generation quality under both RL fine-tuning and Test-time Scaling. Ultimately, our work establishes scalability as a foundational principle for visual RMs, paving the way for more powerful and robust reward modeling in the future of visual generation.

\clearpage

\bibliographystyle{plainnat}
\bibliography{main}

\begin{thebibliography}{72}
\providecommand{\natexlab}[1]{#1}
\providecommand{\url}[1]{\texttt{#1}}
\expandafter\ifx\csname urlstyle\endcsname\relax
  \providecommand{\doi}[1]{doi: #1}\else
  \providecommand{\doi}{doi: \begingroup \urlstyle{rm}\Url}\fi

\bibitem[Bao et~al.(2024)Bao, Xiang, Yue, He, Zhu, Zheng, Zhao, Liu, Wang, and Zhu]{bao2024vidu}
Fan Bao, Chendong Xiang, Gang Yue, Guande He, Hongzhou Zhu, Kaiwen Zheng, Min Zhao, Shilong Liu, Yaole Wang, and Jun Zhu.
\newblock Vidu: a highly consistent, dynamic and skilled text-to-video generator with diffusion models.
\newblock \emph{arXiv preprint arXiv:2405.04233}, 2024.

\bibitem[Betker et~al.(2023)Betker, Goh, Jing, Brooks, Wang, Li, Ouyang, Zhuang, Lee, Guo, et~al.]{betker2023improving}
James Betker, Gabriel Goh, Li~Jing, Tim Brooks, Jianfeng Wang, Linjie Li, Long Ouyang, Juntang Zhuang, Joyce Lee, Yufei Guo, et~al.
\newblock Improving image generation with better captions.
\newblock \emph{Computer Science. https://cdn. openai. com/papers/dall-e-3. pdf}, 2\penalty0 (3):\penalty0 8, 2023.

\bibitem[Black et~al.(2023)Black, Janner, Du, Kostrikov, and Levine]{black2023training}
Kevin Black, Michael Janner, Yilun Du, Ilya Kostrikov, and Sergey Levine.
\newblock Training diffusion models with reinforcement learning.
\newblock \emph{arXiv preprint arXiv:2305.13301}, 2023.

\bibitem[Blattmann et~al.(2023)Blattmann, Dockhorn, Kulal, Mendelevitch, Kilian, Lorenz, Levi, English, Voleti, Letts, et~al.]{blattmann2023stable}
Andreas Blattmann, Tim Dockhorn, Sumith Kulal, Daniel Mendelevitch, Maciej Kilian, Dominik Lorenz, Yam Levi, Zion English, Vikram Voleti, Adam Letts, et~al.
\newblock Stable video diffusion: Scaling latent video diffusion models to large datasets.
\newblock \emph{arXiv preprint arXiv:2311.15127}, 2023.

\bibitem[Bradley and Terry(1952)]{bradley1952rank}
Ralph~Allan Bradley and Milton~E Terry.
\newblock Rank analysis of incomplete block designs: I. the method of paired comparisons.
\newblock \emph{Biometrika}, 39\penalty0 (3/4):\penalty0 324--345, 1952.

\bibitem[Brooks et~al.(2024)Brooks, Peebles, Holmes, DePue, Guo, Jing, Schnurr, Taylor, Luhman, Luhman, et~al.]{brooks2024video}
Tim Brooks, Bill Peebles, Connor Holmes, Will DePue, Yufei Guo, Li~Jing, David Schnurr, Joe Taylor, Troy Luhman, Eric Luhman, et~al.
\newblock Video generation models as world simulators.
\newblock \emph{OpenAI Blog}, 1\penalty0 (8):\penalty0 1, 2024.

\bibitem[Chen et~al.(2023{\natexlab{a}})Chen, Yu, Ge, Yao, Xie, Wu, Wang, Kwok, Luo, Lu, et~al.]{chen2023pixart}
Junsong Chen, Jincheng Yu, Chongjian Ge, Lewei Yao, Enze Xie, Yue Wu, Zhongdao Wang, James Kwok, Ping Luo, Huchuan Lu, et~al.
\newblock Pixart-$\alpha$: Fast training of diffusion transformer for photorealistic text-to-image synthesis.
\newblock \emph{arXiv preprint arXiv:2310.00426}, 2023{\natexlab{a}}.

\bibitem[Chen et~al.(2023{\natexlab{b}})Chen, Wu, Xie, Wu, Li, Xia, Xiao, and Lin]{chen2023control}
Weifeng Chen, Jie Wu, Pan Xie, Hefeng Wu, Jiashi Li, Xin Xia, Xuefeng Xiao, and Liang Lin.
\newblock Control-a-video: Controllable text-to-video generation with diffusion models.
\newblock \emph{CoRR}, 2023{\natexlab{b}}.

\bibitem[Chen et~al.(2024{\natexlab{a}})Chen, Zhu, Sun, Chen, Zhang, and Shen]{chen2024accuracy}
Yanjun Chen, Dawei Zhu, Yirong Sun, Xinghao Chen, Wei Zhang, and Xiaoyu Shen.
\newblock The accuracy paradox in rlhf: When better reward models don't yield better language models.
\newblock \emph{arXiv preprint arXiv:2410.06554}, 2024{\natexlab{a}}.

\bibitem[Chen et~al.(2024{\natexlab{b}})Chen, Wang, Cao, Liu, Gao, Cui, Zhu, Ye, Tian, Liu, et~al.]{chen2024expanding}
Zhe Chen, Weiyun Wang, Yue Cao, Yangzhou Liu, Zhangwei Gao, Erfei Cui, Jinguo Zhu, Shenglong Ye, Hao Tian, Zhaoyang Liu, et~al.
\newblock Expanding performance boundaries of open-source multimodal models with model, data, and test-time scaling.
\newblock \emph{arXiv preprint arXiv:2412.05271}, 2024{\natexlab{b}}.

\bibitem[Dong et~al.(2023)Dong, Xiong, Goyal, Zhang, Chow, Pan, Diao, Zhang, Shum, and Zhang]{dong2023raft}
Hanze Dong, Wei Xiong, Deepanshu Goyal, Yihan Zhang, Winnie Chow, Rui Pan, Shizhe Diao, Jipeng Zhang, Kashun Shum, and Tong Zhang.
\newblock Raft: Reward ranked finetuning for generative foundation model alignment.
\newblock \emph{arXiv preprint arXiv:2304.06767}, 2023.

\bibitem[Esser et~al.(2024)Esser, Kulal, Blattmann, Entezari, M{\"u}ller, Saini, Levi, Lorenz, Sauer, Boesel, et~al.]{esser2024scaling}
Patrick Esser, Sumith Kulal, Andreas Blattmann, Rahim Entezari, Jonas M{\"u}ller, Harry Saini, Yam Levi, Dominik Lorenz, Axel Sauer, Frederic Boesel, et~al.
\newblock Scaling rectified flow transformers for high-resolution image synthesis.
\newblock In \emph{Forty-first international conference on machine learning}, 2024.

\bibitem[Fan et~al.(2023)Fan, Watkins, Du, Liu, Ryu, Boutilier, Abbeel, Ghavamzadeh, Lee, and Lee]{fan2023reinforcement}
Ying Fan, Olivia Watkins, Yuqing Du, Hao Liu, Moonkyung Ryu, Craig Boutilier, Pieter Abbeel, Mohammad Ghavamzadeh, Kangwook Lee, and Kimin Lee.
\newblock Reinforcement learning for fine-tuning text-to-image diffusion models.
\newblock In \emph{Thirty-seventh Conference on Neural Information Processing Systems (NeurIPS) 2023}. Neural Information Processing Systems Foundation, 2023.

\bibitem[Gao et~al.(2025{\natexlab{a}})Gao, Gong, Guo, Hou, Lai, Li, Li, Lian, Liao, Liu, et~al.]{gao2025seedream}
Yu~Gao, Lixue Gong, Qiushan Guo, Xiaoxia Hou, Zhichao Lai, Fanshi Li, Liang Li, Xiaochen Lian, Chao Liao, Liyang Liu, et~al.
\newblock Seedream 3.0 technical report.
\newblock \emph{arXiv preprint arXiv:2504.11346}, 2025{\natexlab{a}}.

\bibitem[Gao et~al.(2025{\natexlab{b}})Gao, Guo, Hoang, Huang, Jiang, Kong, Li, Li, Li, Li, et~al.]{gao2025seedance}
Yu~Gao, Haoyuan Guo, Tuyen Hoang, Weilin Huang, Lu~Jiang, Fangyuan Kong, Huixia Li, Jiashi Li, Liang Li, Xiaojie Li, et~al.
\newblock Seedance 1.0: Exploring the boundaries of video generation models.
\newblock \emph{arXiv preprint arXiv:2506.09113}, 2025{\natexlab{b}}.

\bibitem[Gong et~al.(2025)Gong, Hou, Li, Li, Lian, Liu, Liu, Liu, Lu, Shi, et~al.]{gong2025seedream}
Lixue Gong, Xiaoxia Hou, Fanshi Li, Liang Li, Xiaochen Lian, Fei Liu, Liyang Liu, Wei Liu, Wei Lu, Yichun Shi, et~al.
\newblock Seedream 2.0: A native chinese-english bilingual image generation foundation model.
\newblock \emph{arXiv preprint arXiv:2503.07703}, 2025.

\bibitem[Google(2025)]{veo}
Google.
\newblock Veo.
\newblock \url{https://deepmind.google/models/veo/}, 2025.

\bibitem[Gu et~al.(2024)Gu, Li, Zhang, Chen, Wen, Luo, and Zhu]{multireward}
Xin Gu, Ming Li, Libo Zhang, Fan Chen, Longyin Wen, Tiejian Luo, and Sijie Zhu.
\newblock Multi-reward as condition for instruction-based image editing.
\newblock \emph{arXiv preprint arXiv:2411.04713}, 2024.

\bibitem[Guo et~al.(2023)Guo, Yang, Rao, Liang, Wang, Qiao, Agrawala, Lin, and Dai]{guo2023animatediff}
Yuwei Guo, Ceyuan Yang, Anyi Rao, Zhengyang Liang, Yaohui Wang, Yu~Qiao, Maneesh Agrawala, Dahua Lin, and Bo~Dai.
\newblock Animatediff: Animate your personalized text-to-image diffusion models without specific tuning.
\newblock \emph{arXiv preprint arXiv:2307.04725}, 2023.

\bibitem[Gupta et~al.(2025)Gupta, Ahuja, Lin, Roy, Oosterhuis, de~Rijke, and Shukla]{gupta2025simple}
Shashank Gupta, Chaitanya Ahuja, Tsung-Yu Lin, Sreya~Dutta Roy, Harrie Oosterhuis, Maarten de~Rijke, and Satya~Narayan Shukla.
\newblock A simple and effective reinforcement learning method for text-to-image diffusion fine-tuning.
\newblock \emph{arXiv preprint arXiv:2503.00897}, 2025.

\bibitem[Ho et~al.(2020)Ho, Jain, and Abbeel]{ho2020denoising}
Jonathan Ho, Ajay Jain, and Pieter Abbeel.
\newblock Denoising diffusion probabilistic models.
\newblock \emph{Advances in neural information processing systems}, 33:\penalty0 6840--6851, 2020.

\bibitem[Ideogram(2024)]{Ideogram}
Ideogram.
\newblock Ideogram.
\newblock \url{https://about.ideogram.ai/1.0.}, 2024.

\bibitem[Kirstain et~al.(2023)Kirstain, Polyak, Singer, Matiana, Penna, and Levy]{kirstain2023pick}
Yuval Kirstain, Adam Polyak, Uriel Singer, Shahbuland Matiana, Joe Penna, and Omer Levy.
\newblock Pick-a-pic: An open dataset of user preferences for text-to-image generation.
\newblock \emph{Advances in neural information processing systems}, 36:\penalty0 36652--36663, 2023.

\bibitem[klingai(2025)]{klingai}
klingai.
\newblock klingai.
\newblock \url{https://app.klingai.com/cn/}, 2025.

\bibitem[Kong et~al.(2024)Kong, Tian, Zhang, Min, Dai, Zhou, Xiong, Li, Wu, Zhang, et~al.]{kong2024hunyuanvideo}
Weijie Kong, Qi~Tian, Zijian Zhang, Rox Min, Zuozhuo Dai, Jin Zhou, Jiangfeng Xiong, Xin Li, Bo~Wu, Jianwei Zhang, et~al.
\newblock Hunyuanvideo: A systematic framework for large video generative models.
\newblock \emph{arXiv preprint arXiv:2412.03603}, 2024.

\bibitem[Labs(2024{\natexlab{a}})]{blackforestlabs_flux}
Black~Forest Labs.
\newblock Flux: Official inference repository for flux.1 models, 2024{\natexlab{a}}.
\newblock URL \url{https://github.com/black-forest-labs/flux}.
\newblock Accessed: 2024-11-12.

\bibitem[Labs(2024{\natexlab{b}})]{flux2024}
Black~Forest Labs.
\newblock Flux.
\newblock \url{https://github.com/black-forest-labs/flux}, 2024{\natexlab{b}}.

\bibitem[Li et~al.(2024{\natexlab{a}})Li, Yang, Kuang, Wu, Wang, Xiao, and Chen]{controlnet++}
Ming Li, Taojiannan Yang, Huafeng Kuang, Jie Wu, Zhaoning Wang, Xuefeng Xiao, and Chen Chen.
\newblock Controlnet++: Improving conditional controls with efficient consistency feedback: Project page: liming-ai. github. io/controlnet\_plus\_plus.
\newblock In \emph{European Conference on Computer Vision}, pages 129--147. Springer, 2024{\natexlab{a}}.

\bibitem[Li et~al.(2025)Li, Gu, Chen, Xing, Wen, Chen, and Zhu]{superedit}
Ming Li, Xin Gu, Fan Chen, Xiaoying Xing, Longyin Wen, Chen Chen, and Sijie Zhu.
\newblock Superedit: Rectifying and facilitating supervision for instruction-based image editing.
\newblock \emph{arXiv preprint arXiv:2505.02370}, 2025.

\bibitem[Li et~al.(2024{\natexlab{b}})Li, Koh, and Du]{li2024exploring}
Siting Li, Pang~Wei Koh, and Simon~Shaolei Du.
\newblock Exploring how generative mllms perceive more than clip with the same vision encoder.
\newblock \emph{arXiv preprint arXiv:2411.05195}, 2024{\natexlab{b}}.

\bibitem[Li et~al.(2023)Li, Wang, and Xie]{li2023inverse}
Xianhang Li, Zeyu Wang, and Cihang Xie.
\newblock An inverse scaling law for clip training.
\newblock \emph{Advances in Neural Information Processing Systems}, 36:\penalty0 49068--49087, 2023.

\bibitem[Liao et~al.(2025)Liao, Liu, Wang, Luo, Zhang, Zhao, Wu, Li, Tian, and Huang]{liao2025mogao}
Chao Liao, Liyang Liu, Xun Wang, Zhengxiong Luo, Xinyu Zhang, Wenliang Zhao, Jie Wu, Liang Li, Zhi Tian, and Weilin Huang.
\newblock Mogao: An omni foundation model for interleaved multi-modal generation.
\newblock \emph{arXiv preprint arXiv:2505.05472}, 2025.

\bibitem[Lipman et~al.(2022)Lipman, Chen, Ben-Hamu, Nickel, and Le]{lipman2022flow}
Yaron Lipman, Ricky~TQ Chen, Heli Ben-Hamu, Maximilian Nickel, and Matt Le.
\newblock Flow matching for generative modeling.
\newblock \emph{arXiv preprint arXiv:2210.02747}, 2022.

\bibitem[Liu et~al.(2025{\natexlab{a}})Liu, Liu, Liang, Li, Liu, Wang, Wan, Zhang, and Ouyang]{liu2025flow}
Jie Liu, Gongye Liu, Jiajun Liang, Yangguang Li, Jiaheng Liu, Xintao Wang, Pengfei Wan, Di~Zhang, and Wanli Ouyang.
\newblock Flow-grpo: Training flow matching models via online rl.
\newblock \emph{arXiv preprint arXiv:2505.05470}, 2025{\natexlab{a}}.

\bibitem[Liu et~al.(2025{\natexlab{b}})Liu, Liu, Liang, Yuan, Liu, Zheng, Wu, Wang, Qin, Xia, et~al.]{liu2025improving}
Jie Liu, Gongye Liu, Jiajun Liang, Ziyang Yuan, Xiaokun Liu, Mingwu Zheng, Xiele Wu, Qiulin Wang, Wenyu Qin, Menghan Xia, et~al.
\newblock Improving video generation with human feedback.
\newblock \emph{arXiv preprint arXiv:2501.13918}, 2025{\natexlab{b}}.

\bibitem[Liu et~al.(2025{\natexlab{c}})Liu, Wang, Xu, Ma, Ruan, Li, Liu, and Wu]{liu2025inference}
Zijun Liu, Peiyi Wang, Runxin Xu, Shirong Ma, Chong Ruan, Peng Li, Yang Liu, and Yu~Wu.
\newblock Inference-time scaling for generalist reward modeling.
\newblock \emph{arXiv preprint arXiv:2504.02495}, 2025{\natexlab{c}}.

\bibitem[lumalabs(2024)]{lumalabs}
lumalabs.
\newblock lumalabs.
\newblock \url{https://lumalabs.ai/}, 2024.

\bibitem[Ma et~al.(2025{\natexlab{a}})Ma, Huang, Yan, Chen, Duan, Yin, Wan, Ming, Song, Chen, et~al.]{ma2025step}
Guoqing Ma, Haoyang Huang, Kun Yan, Liangyu Chen, Nan Duan, Shengming Yin, Changyi Wan, Ranchen Ming, Xiaoniu Song, Xing Chen, et~al.
\newblock Step-video-t2v technical report: The practice, challenges, and future of video foundation model.
\newblock \emph{arXiv preprint arXiv:2502.10248}, 2025{\natexlab{a}}.

\bibitem[Ma et~al.(2025{\natexlab{b}})Ma, Tong, Jia, Hu, Su, Zhang, Yang, Li, Jaakkola, Jia, et~al.]{ma2025inference}
Nanye Ma, Shangyuan Tong, Haolin Jia, Hexiang Hu, Yu-Chuan Su, Mingda Zhang, Xuan Yang, Yandong Li, Tommi Jaakkola, Xuhui Jia, et~al.
\newblock Inference-time scaling for diffusion models beyond scaling denoising steps.
\newblock \emph{arXiv preprint arXiv:2501.09732}, 2025{\natexlab{b}}.

\bibitem[Ma et~al.(2025{\natexlab{c}})Ma, Wu, Sun, and Li]{ma2025hpsv3}
Yuhang Ma, Xiaoshi Wu, Keqiang Sun, and Hongsheng Li.
\newblock Hpsv3: Towards wide-spectrum human preference score.
\newblock \emph{arXiv preprint arXiv:2508.03789}, 2025{\natexlab{c}}.

\bibitem[midjourney(2024)]{midjourney}
midjourney.
\newblock midjourney.
\newblock \url{https://www.midjourney.com/home}, 2024.

\bibitem[Oshima et~al.(2025)Oshima, Suzuki, Matsuo, and Furuta]{oshima2025inference}
Yuta Oshima, Masahiro Suzuki, Yutaka Matsuo, and Hiroki Furuta.
\newblock Inference-time text-to-video alignment with diffusion latent beam search.
\newblock \emph{arXiv preprint arXiv:2501.19252}, 2025.

\bibitem[Ouyang et~al.(2022)Ouyang, Wu, Jiang, Almeida, Wainwright, Mishkin, Zhang, Agarwal, Slama, Ray, et~al.]{ouyang2022training}
Long Ouyang, Jeffrey Wu, Xu~Jiang, Diogo Almeida, Carroll Wainwright, Pamela Mishkin, Chong Zhang, Sandhini Agarwal, Katarina Slama, Alex Ray, et~al.
\newblock Training language models to follow instructions with human feedback.
\newblock \emph{Advances in neural information processing systems}, 35:\penalty0 27730--27744, 2022.

\bibitem[Podell et~al.(2023)Podell, English, Lacey, Blattmann, Dockhorn, M{\"u}ller, Penna, and Rombach]{podell2023sdxl}
Dustin Podell, Zion English, Kyle Lacey, Andreas Blattmann, Tim Dockhorn, Jonas M{\"u}ller, Joe Penna, and Robin Rombach.
\newblock Sdxl: Improving latent diffusion models for high-resolution image synthesis.
\newblock \emph{arXiv preprint arXiv:2307.01952}, 2023.

\bibitem[Ramesh et~al.(2022)Ramesh, Dhariwal, Nichol, Chu, and Chen]{ramesh2022hierarchical}
Aditya Ramesh, Prafulla Dhariwal, Alex Nichol, Casey Chu, and Mark Chen.
\newblock Hierarchical text-conditional image generation with clip latents.
\newblock \emph{arXiv preprint arXiv:2204.06125}, 1\penalty0 (2):\penalty0 3, 2022.

\bibitem[Razin et~al.(2025)Razin, Wang, Strauss, Wei, Lee, and Arora]{razin2025makes}
Noam Razin, Zixuan Wang, Hubert Strauss, Stanley Wei, Jason~D Lee, and Sanjeev Arora.
\newblock What makes a reward model a good teacher? an optimization perspective.
\newblock \emph{arXiv preprint arXiv:2503.15477}, 2025.

\bibitem[recraft(2024)]{recraft}
recraft.
\newblock recraft.
\newblock \url{https://www.recraft.ai/}, 2024.

\bibitem[Ren et~al.(2024)Ren, Wu, Lu, Kuang, Xia, Wang, Wang, Zhu, Xie, Wang, et~al.]{ren2024byteedit}
Yuxi Ren, Jie Wu, Yanzuo Lu, Huafeng Kuang, Xin Xia, Xionghui Wang, Qianqian Wang, Yixing Zhu, Pan Xie, Shiyin Wang, et~al.
\newblock Byteedit: Boost, comply and accelerate generative image editing.
\newblock In \emph{European Conference on Computer Vision}, pages 184--200. Springer, 2024.

\bibitem[Rombach et~al.(2022)Rombach, Blattmann, Lorenz, Esser, and Ommer]{rombach2022high}
Robin Rombach, Andreas Blattmann, Dominik Lorenz, Patrick Esser, and Bj{\"o}rn Ommer.
\newblock High-resolution image synthesis with latent diffusion models.
\newblock In \emph{Proceedings of the IEEE/CVF conference on computer vision and pattern recognition}, pages 10684--10695, 2022.

\bibitem[Runway(2025)]{runwaygen4}
Runway.
\newblock Runway.
\newblock \url{https://runwayml.com/research/introducing-runway-gen-4}, 2025.

\bibitem[Seawead et~al.(2025)Seawead, Yang, Lin, Zhao, Lin, Ma, Guo, Chen, Qi, Wang, et~al.]{seawead2025seaweed}
Team Seawead, Ceyuan Yang, Zhijie Lin, Yang Zhao, Shanchuan Lin, Zhibei Ma, Haoyuan Guo, Hao Chen, Lu~Qi, Sen Wang, et~al.
\newblock Seaweed-7b: Cost-effective training of video generation foundation model.
\newblock \emph{arXiv preprint arXiv:2504.08685}, 2025.

\bibitem[Sohl-Dickstein et~al.(2015)Sohl-Dickstein, Weiss, Maheswaranathan, and Ganguli]{sohl2015deep}
Jascha Sohl-Dickstein, Eric Weiss, Niru Maheswaranathan, and Surya Ganguli.
\newblock Deep unsupervised learning using nonequilibrium thermodynamics.
\newblock In \emph{International conference on machine learning}, pages 2256--2265. pmlr, 2015.

\bibitem[Song et~al.(2020)Song, Sohl-Dickstein, Kingma, Kumar, Ermon, and Poole]{song2020score}
Yang Song, Jascha Sohl-Dickstein, Diederik~P Kingma, Abhishek Kumar, Stefano Ermon, and Ben Poole.
\newblock Score-based generative modeling through stochastic differential equations.
\newblock \emph{arXiv preprint arXiv:2011.13456}, 2020.

\bibitem[Sun et~al.(2024)Sun, Jiang, Chen, Zhang, Peng, Luo, and Yuan]{sun2024autoregressive}
Peize Sun, Yi~Jiang, Shoufa Chen, Shilong Zhang, Bingyue Peng, Ping Luo, and Zehuan Yuan.
\newblock Autoregressive model beats diffusion: Llama for scalable image generation.
\newblock \emph{arXiv preprint arXiv:2406.06525}, 2024.

\bibitem[Wallace et~al.(2024)Wallace, Dang, Rafailov, Zhou, Lou, Purushwalkam, Ermon, Xiong, Joty, and Naik]{wallace2024diffusion}
Bram Wallace, Meihua Dang, Rafael Rafailov, Linqi Zhou, Aaron Lou, Senthil Purushwalkam, Stefano Ermon, Caiming Xiong, Shafiq Joty, and Nikhil Naik.
\newblock Diffusion model alignment using direct preference optimization.
\newblock In \emph{Proceedings of the IEEE/CVF Conference on Computer Vision and Pattern Recognition}, pages 8228--8238, 2024.

\bibitem[Wan et~al.(2025)Wan, Wang, Ai, Wen, Mao, Xie, Chen, Yu, Zhao, Yang, et~al.]{wan2025wan}
Team Wan, Ang Wang, Baole Ai, Bin Wen, Chaojie Mao, Chen-Wei Xie, Di~Chen, Feiwu Yu, Haiming Zhao, Jianxiao Yang, et~al.
\newblock Wan: Open and advanced large-scale video generative models.
\newblock \emph{arXiv preprint arXiv:2503.20314}, 2025.

\bibitem[Wang et~al.(2025{\natexlab{a}})Wang, Lin, Lu, Yu, Zhang, Huang, Zheng, Dang, Fan, Ren, et~al.]{wang2025worldpm}
Binghai Wang, Runji Lin, Keming Lu, Le~Yu, Zhenru Zhang, Fei Huang, Chujie Zheng, Kai Dang, Yang Fan, Xingzhang Ren, et~al.
\newblock Worldpm: Scaling human preference modeling.
\newblock \emph{arXiv preprint arXiv:2505.10527}, 2025{\natexlab{a}}.

\bibitem[Wang et~al.(2024)Wang, Zhang, Luo, Sun, Cui, Wang, Zhang, Wang, Li, Yu, et~al.]{wang2024emu3}
Xinlong Wang, Xiaosong Zhang, Zhengxiong Luo, Quan Sun, Yufeng Cui, Jinsheng Wang, Fan Zhang, Yueze Wang, Zhen Li, Qiying Yu, et~al.
\newblock Emu3: Next-token prediction is all you need.
\newblock \emph{arXiv preprint arXiv:2409.18869}, 2024.

\bibitem[Wang et~al.(2025{\natexlab{b}})Wang, Li, Zang, Wang, Lu, Jin, and Wang]{wang2025unified}
Yibin Wang, Zhimin Li, Yuhang Zang, Chunyu Wang, Qinglin Lu, Cheng Jin, and Jiaqi Wang.
\newblock Unified multimodal chain-of-thought reward model through reinforcement fine-tuning.
\newblock \emph{arXiv preprint arXiv:2505.03318}, 2025{\natexlab{b}}.

\bibitem[Wen et~al.(2024)Wen, Lou, Lu, Lin, Yu, Lu, He, Han, Zhang, and Sun]{wen2024rethinking}
Xueru Wen, Jie Lou, Yaojie Lu, Hongyu Lin, Xing Yu, Xinyu Lu, Ben He, Xianpei Han, Debing Zhang, and Le~Sun.
\newblock Rethinking reward model evaluation: Are we barking up the wrong tree?
\newblock \emph{arXiv preprint arXiv:2410.05584}, 2024.

\bibitem[Wu et~al.(2025)Wu, Li, Zhou, Lin, Gao, Yan, Yin, Bai, Xu, Chen, et~al.]{wu2025qwen}
Chenfei Wu, Jiahao Li, Jingren Zhou, Junyang Lin, Kaiyuan Gao, Kun Yan, Sheng-ming Yin, Shuai Bai, Xiao Xu, Yilei Chen, et~al.
\newblock Qwen-image technical report.
\newblock \emph{arXiv preprint arXiv:2508.02324}, 2025.

\bibitem[Wu et~al.(2023{\natexlab{a}})Wu, Sun, Zhu, Zhao, and Li]{wu2023better}
Xiaoshi Wu, Keqiang Sun, Feng Zhu, Rui Zhao, and Hongsheng Li.
\newblock Better aligning text-to-image models with human preference.
\newblock \emph{arXiv preprint arXiv:2303.14420}, 1\penalty0 (3), 2023{\natexlab{a}}.

\bibitem[Wu et~al.(2023{\natexlab{b}})Wu, Sun, Zhu, Zhao, and Li]{wu2023human}
Xiaoshi Wu, Keqiang Sun, Feng Zhu, Rui Zhao, and Hongsheng Li.
\newblock Human preference score: Better aligning text-to-image models with human preference.
\newblock In \emph{Proceedings of the IEEE/CVF International Conference on Computer Vision}, pages 2096--2105, 2023{\natexlab{b}}.

\bibitem[Xu et~al.(2023)Xu, Liu, Wu, Tong, Li, Ding, Tang, and Dong]{xu2023imagereward}
Jiazheng Xu, Xiao Liu, Yuchen Wu, Yuxuan Tong, Qinkai Li, Ming Ding, Jie Tang, and Yuxiao Dong.
\newblock Imagereward: Learning and evaluating human preferences for text-to-image generation.
\newblock \emph{Advances in Neural Information Processing Systems}, 36:\penalty0 15903--15935, 2023.

\bibitem[Xu et~al.(2024)Xu, Huang, Cheng, Yang, Xu, Wang, Duan, Yang, Jin, Li, et~al.]{xu2024visionreward}
Jiazheng Xu, Yu~Huang, Jiale Cheng, Yuanming Yang, Jiajun Xu, Yuan Wang, Wenbo Duan, Shen Yang, Qunlin Jin, Shurun Li, et~al.
\newblock Visionreward: Fine-grained multi-dimensional human preference learning for image and video generation.
\newblock \emph{arXiv preprint arXiv:2412.21059}, 2024.

\bibitem[Xu et~al.(2025)Xu, Zuo, Xin, Yue, Yan, and Wu]{xu2025unified}
Wenyuan Xu, Xiaochen Zuo, Chao Xin, Yu~Yue, Lin Yan, and Yonghui Wu.
\newblock A unified pairwise framework for rlhf: Bridging generative reward modeling and policy optimization.
\newblock \emph{arXiv preprint arXiv:2504.04950}, 2025.

\bibitem[Xue et~al.(2025)Xue, Wu, Gao, Kong, Zhu, Chen, Liu, Liu, Guo, Huang, et~al.]{xue2025dancegrpo}
Zeyue Xue, Jie Wu, Yu~Gao, Fangyuan Kong, Lingting Zhu, Mengzhao Chen, Zhiheng Liu, Wei Liu, Qiushan Guo, Weilin Huang, et~al.
\newblock Dancegrpo: Unleashing grpo on visual generation.
\newblock \emph{arXiv preprint arXiv:2505.07818}, 2025.

\bibitem[Ye et~al.(2025)Ye, Chen, Li, Huang, Luo, and Qi]{ye2025schedule}
Zilyu Ye, Zhiyang Chen, Tiancheng Li, Zemin Huang, Weijian Luo, and Guo-Jun Qi.
\newblock Schedule on the fly: Diffusion time prediction for faster and better image generation.
\newblock In \emph{Proceedings of the Computer Vision and Pattern Recognition Conference}, pages 23412--23422, 2025.

\bibitem[Zeng et~al.(2024)Zeng, Wei, Zheng, Zou, Wei, Zhang, and Li]{zeng2024make}
Yan Zeng, Guoqiang Wei, Jiani Zheng, Jiaxin Zou, Yang Wei, Yuchen Zhang, and Hang Li.
\newblock Make pixels dance: High-dynamic video generation.
\newblock In \emph{Proceedings of the IEEE/CVF Conference on Computer Vision and Pattern Recognition}, pages 8850--8860, 2024.

\bibitem[Zhang et~al.(2024{\natexlab{a}})Zhang, Wu, Chen, Ji, Xiao, Huang, and Han]{zhang2024onlinevpo}
Jiacheng Zhang, Jie Wu, Weifeng Chen, Yatai Ji, Xuefeng Xiao, Weilin Huang, and Kai Han.
\newblock Onlinevpo: Align video diffusion model with online video-centric preference optimization.
\newblock \emph{arXiv preprint arXiv:2412.15159}, 2024{\natexlab{a}}.

\bibitem[Zhang et~al.(2024{\natexlab{b}})Zhang, Wu, Ren, Xia, Kuang, Xie, Li, Xiao, Huang, Wen, et~al.]{zhang2024unifl}
Jiacheng Zhang, Jie Wu, Yuxi Ren, Xin Xia, Huafeng Kuang, Pan Xie, Jiashi Li, Xuefeng Xiao, Weilin Huang, Shilei Wen, et~al.
\newblock Unifl: Improve latent diffusion model via unified feedback learning.
\newblock \emph{Advances in Neural Information Processing Systems}, 37:\penalty0 67355--67382, 2024{\natexlab{b}}.

\bibitem[Ziegler et~al.(2019)Ziegler, Stiennon, Wu, Brown, Radford, Amodei, Christiano, and Irving]{ziegler2019fine}
Daniel~M Ziegler, Nisan Stiennon, Jeffrey Wu, Tom~B Brown, Alec Radford, Dario Amodei, Paul Christiano, and Geoffrey Irving.
\newblock Fine-tuning language models from human preferences.
\newblock \emph{arXiv preprint arXiv:1909.08593}, 2019.

\end{thebibliography}

\clearpage


\end{CJK*}
\end{document}